\documentclass{article}

\PassOptionsToPackage{numbers, compress}{natbib}

\usepackage[preprint]{neurips_2026}

\usepackage[utf8]{inputenc}
\usepackage[T1]{fontenc}

\usepackage{microtype}
\usepackage{graphicx}

\usepackage[font=small]{subcaption}
\usepackage{booktabs}
\usepackage{url}
\usepackage{hyperref}
\usepackage{titletoc}

\usepackage{amsmath}
\usepackage{amssymb}
\usepackage{amsfonts}
\usepackage{mathtools}
\usepackage{amsthm}

\usepackage{multirow}
\usepackage{makecell}
\usepackage{xcolor}
\usepackage{nicefrac}

\usepackage{algorithm}

\usepackage{algpseudocode}
\usepackage{float}

\usepackage{enumerate}
\usepackage{enumitem}

\usepackage[capitalize,noabbrev]{cleveref}

\theoremstyle{plain}

\theoremstyle{definition}

\theoremstyle{remark}

\usepackage[disable,textsize=tiny]{todonotes}

\title{DreamPolicy: A Unified World-model Policy for Scalable Humanoid Locomotion}

\author{
  Yahao Fan$^{1,2,*}$ \hspace{1em}
  Tianxiang Gui$^{1,*}$ \hspace{1em}
  Kaiyang Ji$^{1,}$\thanks{Equal contribution. $\dag$ Corresponding author.} \hspace{1em}
  Shutong Ding$^{1}$ \hspace{1em}
  Chixuan Zhang$^{1}$ \\
  \And
  Yifeng Xu$^{1}$ \hspace{1em}
  Ke Yang$^{1}$ \hspace{1em}
  Jiayuan Gu$^{1}$ \hspace{1em}
  Jingyi Yu$^{1}$ \hspace{1em}
  Jingya Wang$^{1,\dag}$ \hspace{1em}
  Ye Shi$^{1,2,\dag}$ \\
  \vspace{1em}\\
  $^1$ShanghaiTech University \hspace{1em} $^2$InstAdapt
  \vspace{1em}\\
  \texttt{\{fanyh12024,guitx2025,jiky2024,dingsht,zhangchx12024\}@shanghaitech.edu.cn} \\
  \texttt{\{2546154803,923526244\}@qq.com} \\
  \texttt{\{gujy1,yujingyi,wangjingya,shiye\}@shanghaitech.edu.cn}
}

\begin{document}

\maketitle

\begin{abstract}
Achieving versatile humanoid locomotion with a single policy presents a critical scalability challenge. Prevailing methods often rely on distilling multiple terrain-specific teacher policies into a unified student policy. However, while such distillation captures basic locomotion primitives, it struggles to organically compose these skills to adapt to complex environments, resulting in poor generalization to novel composite terrains unseen during training. To overcome this, we present DreamPolicy, a unified framework that integrates offline data with a diffusion-based world model, enabling a single policy to master both known and unseen terrains. Central to our approach is a terrain-aware world model, driven by an autoregressive diffusion world model trained on aggregated rollouts from specialized policies.
This model synthesizes physically plausible future trajectories, which serve as dynamic objectives for a conditioned policy, thereby bypassing manual reward engineering. Unlike distillation, our world model captures generalizable locomotion skills, allowing for robust zero-shot transfer to unseen composite terrains. DreamPolicy naturally scales with data availability. As the offline dataset expands, the diffusion world model continuously acquires richer skills. Experiments demonstrate that DreamPolicy outperforms the strongest baseline by up to 27\% on unseen terrains and 38\% on combined terrains. By unifying world model-based planning and policy learning, DreamPolicy breaks the "one task, one policy" bottleneck and establishes a scalable, data-driven paradigm for generalist humanoid control.
\end{abstract}

\section{Introduction}

\begin{figure}
    \centering
    \includegraphics[width=1.0\linewidth]{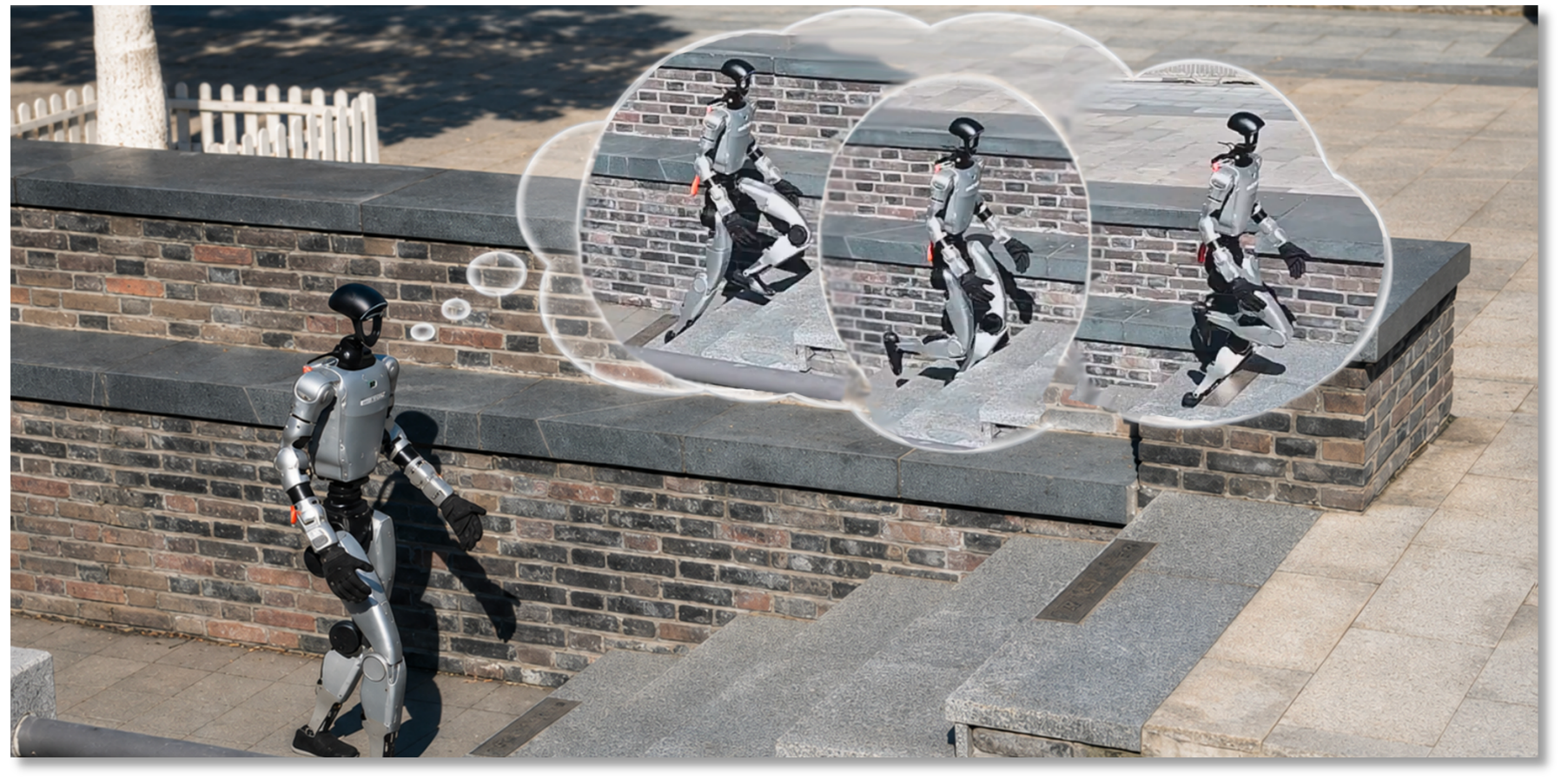}
    \caption{We propose DreamPolicy, a scalable unified policy that leverages a diffusion-based world model to enable versatile humanoid locomotion across diverse terrains.}
    \label{fig:teaser}
\end{figure}

The rapid advancement of embodied intelligence has positioned legged robots as pivotal agents for bridging virtual intelligence with physical-world interactions, particularly in scenarios requiring dynamic balance \citep{vukobratovic1972stability, xie2025humanoid}, dexterous manipulation, and adaptive locomotion. Recent advances in legged locomotion have enabled quadrupedal robots to perform dynamic tasks such as parkour \citep{zhu2026ttt, cheng2024extreme, zhuang2023robot, zargarbashi2024robotkeyframing, fu2021minimizing} and whole body loco-manipulation \cite{fu2022deep, liu2024visual}. 
Humanoid robots, with their anthropomorphic design, hold immense potential for operating in human-centric environments, from locomotion \citep{wang2026x, zhuang2024humanoid, radosavovic2024learning, radosavovic2024humanoid} to manipulation \citep{chen2025rhino, gu2025humanoid}, even whole-body loco-manipulation tasks \citep{cheng2024open, cheng2024expressive, fu2024humanplus, he2024hover, xue2025unified}.  

Notably, the majority of these works struggle with robust performance when facing combinations of complex scenes. 
To better exploit the morphological advantages of humanoid robots and achieve more human-like and natural motion, recent approaches incorporate pre-trained models \citep{zhang2025flam, albaba2025nil} or human motion priors \citep{zhuang2026deep, wu2026perceptive, zhang2024whole, fu2024humanplus, he2024learning, jiang2024harmon} to guide policy learning. These priors reduce the exploration burden in reinforcement learning and encourage the discovery of more feasible action spaces. However, effectively leveraging these priors to handle diverse challenge scenarios (e.g., slopes, gaps, uneven terrain) and external disturbances \citep{jenelten2019dynamic, radosavovic2024humanoid, wang2025beamdojo, zhuang2024humanoid, ren2025vb} remains a significant hurdle. Specifically, most learning-based approaches tend to overfit to specific training conditions \citep{gu2024advancing, long2024learninghumanoidlocomotionperceptive} and rely on task-specific rewards that are difficult to tune for mixed environments. Furthermore, while policy distillation \cite{zhuang2023robot} aggregates specialized skills into a unified policy, it typically compresses diverse behaviors into a single mean representation, struggling to capture the complex, multi-modal distribution of motions required for mixed terrains. Consequently, online distillation methods fail to adapt to distributions outside their curated training sets. Hence, instead of seeking a "one-size-fits-all" policy, designing a data-driven framework capable of organically integrating terrain-conditioned motion priors through scalable data is of exceptional importance.

To address these challenges, we propose \textbf{DreamPolicy}, a novel framework that augments conventional reinforcement learning through the synergistic integration of a terrain-aware autoregressive diffusion world model. Rather than relying on abstract latent representations, our approach utilizes diffusion-synthesized future state predictions as explicit, terrain-adaptive reference trajectories. The framework operates in three distinct stages: First, we aggregate a diverse Humanoid kinematic dataset by training specialized RL policies across five distinct terrains. Subsequently, we train an autoregressive Diffusion World Model on this dataset to learn the cross-scenario state-action distributions, enabling the probabilistic synthesis of physically plausible trajectories. Finally, we optimize a physics-constrained RL policy conditioned on these diffusion-generated references. This architecture effectively decouples reward engineering from environmental complexity, allowing the system to continuously improve through dataset expansion and leverage the diffusion model's capability to generalize across complex terrain distributions. The contributions of our work are summarized as follows: 

\begin{itemize}[leftmargin=4pt, rightmargin=4pt]

\item \textbf{Diffusion-based World Model for Humanoid Locomotion.} We firstly introduce a diffusion-based world model to humanoid locomotion that directly synthesizes multi-step future state trajectories as dynamic guidance for policy learning, which avoids extensive manual reward engineering.

\item \textbf{Terrain-aware Motion Priors for Skill Composition.} 
By training on a multi-terrain dataset of specialized policies, our diffusion world model learns structured, terrain-conditioned motion priors, enabling dynamic skill composition and zero-shot generalization to unseen terrains.

\item \textbf{Superior Performance in Unseen Scenarios.} We show that DreamPolicy benefits consistently from scalable offline data and generalizes beyond the terrains used to train the source policies. 
In comprehensive experiments, DreamPolicy achieves high success rates on training terrains and outperforms the strongest baseline by up to 27\% on unseen terrains and 38\% on combined terrains, demonstrating its effectiveness for robust and generalizable humanoid locomotion.

\end{itemize}

\section{Related Work}

\begin{figure*}[ht]
  \centering
  \includegraphics[width=1.0\textwidth]{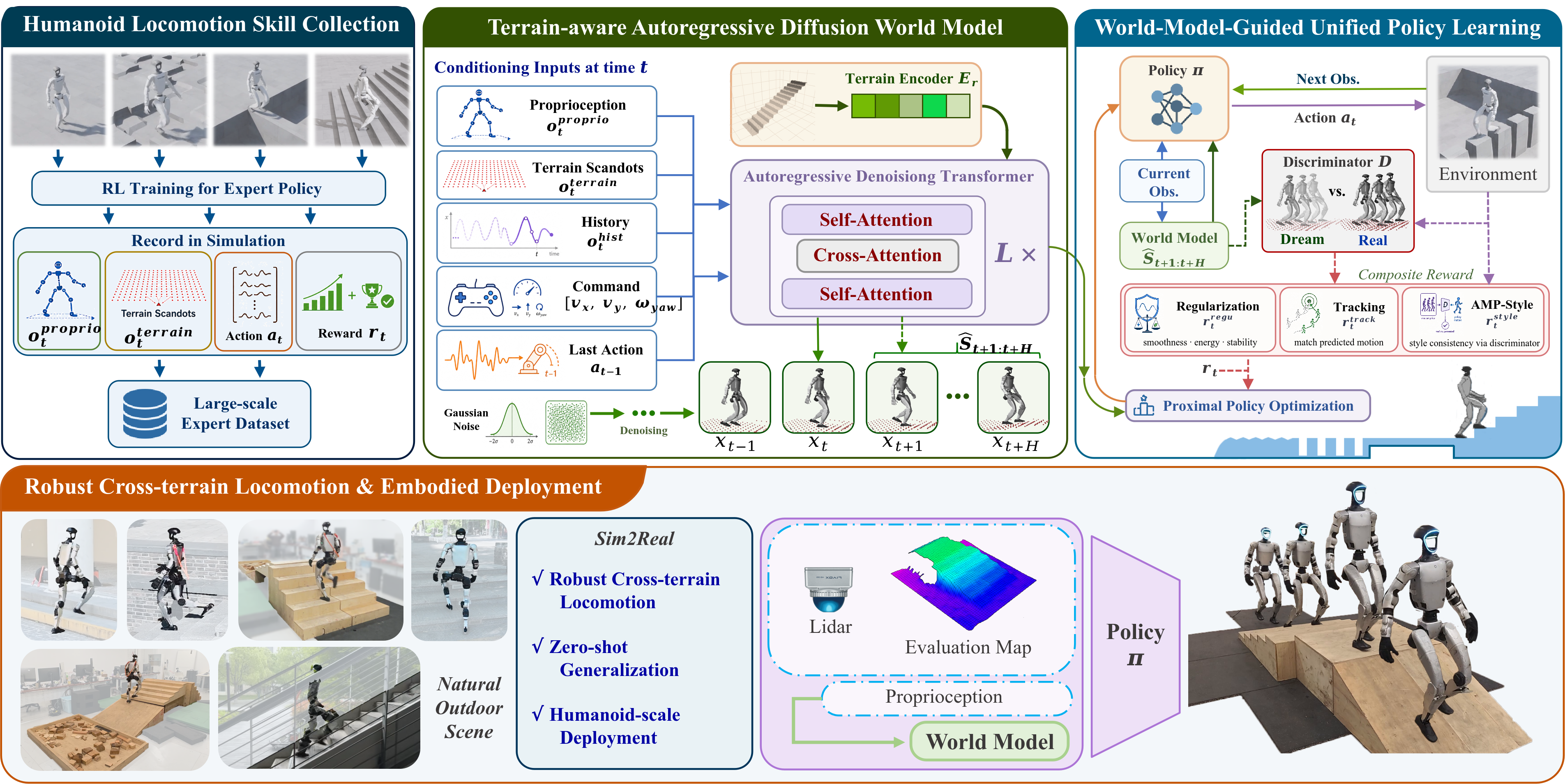}
  \caption{Framework of DreamPolicy. The system is decomposed into two parts: (1) Terrain-aware humanoid locomotion skill data collection, sufficient humanoid locomotion skill data on various challenging terrains will be collected; (2) DreamPolicy, training a unified policy across diverse terrains using skill data and humanoid motion imagery via Auto-Regressive Diffusion World Model.}
  \label{fig:overview}
\end{figure*}

\subsection{Humanoid Locomotion and Generalist Control}
Recent advances in humanoid robotics have predominantly relied on hierarchical reinforcement learning, where methods like Extreme Parkour \cite{cheng2024extreme} and Distillation-PPO \cite{zhang2025distillation} employ specialized skills for robust locomotion. To bridge the Sim2Real gap, works such as DWL \cite{gu2024advancing} and Humanoid Perception Controller \cite{sun2025learningperceptive} focus on latent state reconstruction. However, these approaches often suffer from heavy reliance on manual reward engineering. To address this, data-driven methods like H2O \cite{he2024learning} and OmniH2O \cite{he2024omnih2o} utilize retargeted human motion priors \cite{mahmood2019amass}. 
In contrast, our approach leverages collected native humanoid kinematics, lightens both manual reward design and retargeting process while ensuring physical plausibility.
While foundation models like Octo \cite{team2024octo}, GR00T \cite{bjorck2025gr00t}, and RT-2 \cite{zitkovich2023rt} demonstrate that scaling data significantly boosts generalization, they primarily focus on manipulation or cross-embodiment alignment \cite{bohlinger2024one, huang2020one}. The foundation models of humanoid robots focus more on motion tracking \citep{zeng2025behavior, li2025bfm}.
The closest work \cite{zhuang2023robot} employs distillation but struggles to generalize to novel composite terrains. Our framework bridges this gap by employing an autoregressive diffusion model trained on native kinematics, ensuring robust zero-shot adaptability without the limitations of retargeting or policy distillation.

\subsection{Diffusion Models for Robot Learning}

Beyond their success in vision \citep{song2020score, ho2022video}, diffusion models have become an expressive generative paradigm in robot learning \citep{chi2023diffusion, wangdiffusion, janner2022planning}. Unlike traditional stepwise policies that directly map observations to actions, diffusion-based methods generate actions or trajectories by denoising from a prior, naturally supporting multi-modal behavior modeling and long-horizon temporal consistency \citep{wen2025diffusionvla, liu2025hybridvla}. This property is particularly relevant for locomotion, where different terrains often require distinct but smoothly connected motion modes. DiPPeR \citep{liu2024dipper} utilizes diffusion for terrain-aware path planning, while DiffuseLoco \cite{huangdiffuseloco} and BiRoDiff \cite{mothish2024birodiff} show that diffusion policies can unify diverse gait behaviors and adapt to unseen scenarios. Complementarily, PARC \citep{xu2025parc, zhang2026learning} demonstrates the value of combining generative motion models with physics-based tracking to expand agile terrain-traversal skills. These approaches highlight the potential of diffusion as a scalable motion generator, effectively bridging the gap between high-level planning and low-level control \citep{yu2025discovery, yuan2024preference}. However, existing methods typically use diffusion as an action policy, a path planner, or a reference-motion generator. In contrast, DreamPolicy treats diffusion as a terrain-aware world model that predicts future humanoid state trajectories conditioned on local terrain and recent robot history, turning multi-modal locomotion priors into online goals for unified policy learning.

\subsection{World Models for Humanoid Locomotion}

Explicit predictive world models remain relatively underexplored in humanoid locomotion, but recent work has begun to model humanoid-environment dynamics more directly. Humanoid-specific methods such as DWL \citep{gu2024advancing}, WMR \citep{sun2025learning}, HuWo \citep{zheng2025huwobuilding}, HAIC \citep{li2026haic}, and Ego-VCP \citep{liu2025ego} introduce predictive models for terrain traversal, physical interaction, or contact planning. These methods improve robustness by reconstructing latent dynamics, regularizing policy learning, or supporting future contact reasoning, but the learned model is usually used as an auxiliary representation rather than as an explicit source of execution-time motion guidance. More broadly, classical model-based RL methods such as PlaNet \citep{hafner2019learning}, DreamerV3 \citep{hafner2023mastering}, TD-MPC \citep{hansen2022temporal}, and DayDreamer \citep{wu2023daydreamer} learn compact latent dynamics and improve control through imagined rollouts or latent-space planning. Recent simulator-style world models further scale this idea in robotics and autonomous driving by using action-conditioned predictors for policy optimization or evaluation \citep{huang2026towards, li2025robotic, yang2023unisim}. In contrast, DreamPolicy employs a diffusion world model to generate physically plausible future humanoid motion references online, while a separate low-level controller tracks these references for execution. This explicit separation between high-level generative foresight and low-level control makes the predicted future directly actionable and improves robustness on unseen composite terrains.

\section{Terrain-aware Humanoid Locomotion Skills Collection} \label{method:dataset}
In this section, we introduce our pipeline for collecting terrain-aware humanoid locomotion skill data. Section \ref{subsec:benchmark} outlines the modular terrain benchmarks designed to evaluate policy robustness. In section \ref{subsec:collection}, we detail the training of specialized policies to obtain primitive skills.

\subsection{Humanoid Locomotion Benchmarks with Diverse Terrains} \label{subsec:benchmark}

\begin{figure*}[ht]
  \centering
  \includegraphics[width=0.95\textwidth]{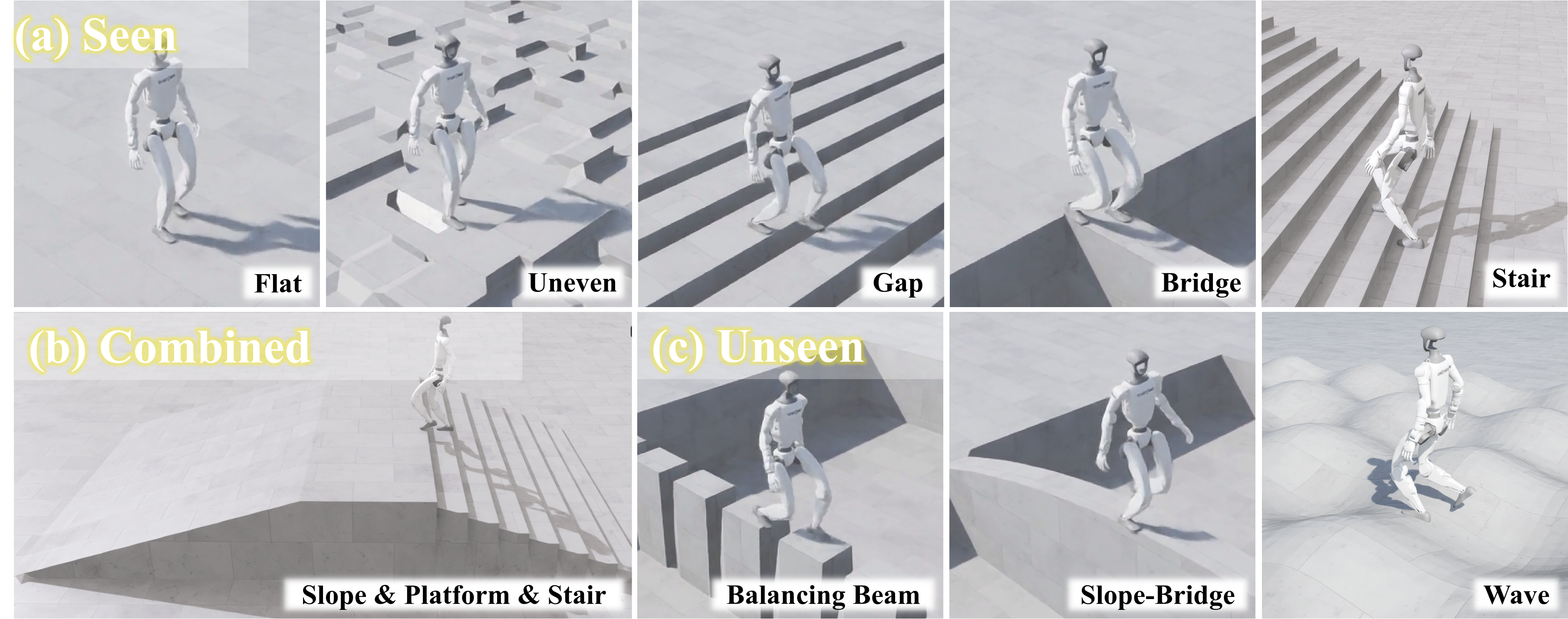}
  \caption{Visualization of the modular terrains. Top row: Primitive terrains used for training specialized expert policies. Bottom row: Combined terrains and unseen terrains used for evaluation.}
  \label{exp:terrain_figure}
\end{figure*}

We developed a modular humanoid terrain benchmark to facilitate skill acquisition and evaluation. This benchmark includes over a dozen specialized primitive terrains (e.g., Gap, Stair, Bridge) for expert policy training, and challenging terrains (e.g., Balancing Beam, Slope-Bridge) for testing zero-shot generalization, as visualized in Figure \ref{exp:terrain_figure}. These terrains can be flexibly combined to generate a wide variety of scenarios, ensuring that the collected data covers a rich distribution of kinematic transitions. Details of the benchmark are provided in the Appendix \ref{sup:benchmark}.

Performance is quantitatively measured using metrics such as success rate and completion rate. 
An efficient, reusable recording module supports comprehensive data collection in simulation, capturing proprioception, exteroceptive data (e.g., sparse scandots), and rewards. Leveraging the parallelized architecture of the simulator, we efficiently acquire a set of terrain-specific policies.
Ultimately, we collected data from five expert policies, yielding a dataset of 75 million transitions.
As shown in Figures \ref{exp:scaling_nums}, this expert dataset was crucial to DreamPolicy’s strong performance.

\subsection{Humanoid Primitive Skills Collection via Specialized Policy Learning} \label{subsec:collection}

To acquire primitive humanoid locomotion skills, we first train multiple physically feasible policies \(\pi = [\pi_1, \pi_2, ..., \pi_n]\) in Isaaclab \cite{mittal2025isaac} using reinforcement learning. Here, \(\pi_i\) represents an expert policy specialized for a single terrain.

While each policy is specialized for individual terrains and lacks generalization capabilities across diverse composite scenarios (as discussed in Section \ref{exp:generalization}), they serve as high-quality primitives.
Despite environmental variations, all policies share the same network architecture to facilitate parameter sharing. The training adopts a two-phase progressive curriculum: First, a foundational locomotion policy is trained from scratch under strict regularization to establish basic walking stability. Second, scenario-specific policies are refined through terrain-specific reward shaping. This separation ensures that while reward engineering is required for the primitives, it is decoupled from the subsequent task of generalized locomotion on novel terrains. Following \cite{zhu2026hiking}, we incorporate edge-aware penalization to discourage risky footholds near terrain boundaries without requiring precise foothold specification.

\textbf{Observation Space and Action Space. }The policy obsevations, which denotes as \(o_t\), consist of:
\begin{equation}
    \label{eqa: observation}
    o_t = [o_t^\text{proprio},  o_t^\text{terrain}, o_t^{\text{hist}}, c_t, a_{t-1}],
\end{equation}
where \(o_t^\text{proprio}\) is the proprioceptive state, \(o_t^\text{terrain}\) denotes exteroceptive information, \(o_t^{\text{hist}}\) represents the observation of the previous timestep, \(c_t\) is the user command, and \(a_{t-1}\) is the action of the last timestep. 
Proprioceptive information is defined as \(o_t^\text{proprio} = [w_t, g_t, \theta_t, \dot{\theta_t}]\), containing base angular velocity, projected gravity vector, joint positions, and joint velocities. 
The command \(c_t = [v_x, v_y, w_{\text{yaw}}]\) includes the desired linear velocities and yaw rate, which are resampled randomly within a predefined range at fixed intervals during training to ensure robustness.
Perceptive information \(o_t^{\text{terrain}}\) consists of sparse scandots sampled from the egocentric height map in the robot's frame, aligning its z-axis to the negative gravity direction. 
The action \(a_t \in \mathbb{R}^{29}\) represents the target position for all joints of the robot. More details about the specialized policies are provided in the Appendix~\ref{sup:specialist}. 

\section{Unified Policy for Cross-Terrain Humanoid Locomotion} \label{method:dreampolicy}

In this section, we detail the implementation of the DreamPolicy framework, as illustrated in Figure \ref{fig:overview}. Section \ref{method:diffusion} introduces the Terrain-Aware Autoregressive Diffusion World Model. Section \ref{sec:gcrl} describes how we formulate the Unified Policy, which integrates diffusion-generated trajectory tracking with an AMP-style \cite{peng2021amp} manifold regularization.

\subsection{Terrain-aware Autoregressive Diffusion World Model} \label{method:diffusion}
Based on the large-scale multi-expert locomotion skill data collected in Section \ref{subsec:collection}, we employ a terrain‐aware autoregressive diffusion world model to predict adaptable humanoid future trajectories.

\textbf{Autoregressive Denoising on State Space.} 
To ensure the model is deployable on real robots without relying on privileged global ground truth, we define the humanoid state $s_t$ consistently with the policy observation space defined in Section \ref{subsec:collection}. Specifically, $s_t$ is a concatenation of proprioceptive and local perceptive features: 
\begin{equation}
    s_t = [o_t^\text{proprio}, c_t, a_{t-1}],
\end{equation}
where $o_t^\text{proprio}$ includes base angular velocity, gravity vector, and joint states; $c_t$ is the user command; and $a_{t-1}$ is the previous action. Crucially, this state space is invariant to global position and yaw, enabling robust generalization across unbounded environments.

Rather than applying diffusion to actions (which are often discrete or vary in dimension), as analysed in \cite{ajay2023is}, we diffuse over the state trajectories. At each control step $t$, given the history of $K$ frames of humanoid state ${\mathbf{x}_{t-1} = [s_{t-H+1}, \dots, s_t]}$ and current local terrain embedding $y_t = \mathcal{E}_{\text{terrain}}(o^{\text{terrain}}_t)$, the goal is to sample the next $H$ humanoid states $\mathbf{x}_t = [s_{t+1}\dots, s_{t+K}]$ conditioned on $(x_{t-1}, y_t)$.

For generative modeling, we follow a Denoising Diffusion Probabilistic Model (DDPM) \cite{ho2020denoising} framework by gradually adding Gaussian noise to the clean humanoid state trajectory $\mathbf{x}_t^0$ over $N$ steps. Formally, the forward process for diffusion timestep $n\in\{1, \dots, N\}$ is:
\begin{equation}
    q(\mathbf{x}^{n} | \mathbf{x}^{n -1}) = \mathcal{N}(\sqrt{\alpha_n} \mathbf{x}^{n-1}, (1 - \alpha_n) \mathbf{I}),
\end{equation}
where $\alpha_n \in (0, 1)$ are fixed variance-scheduling parameter for step $n$. 
In the reverse process, we form the input to the denoiser as $(\mathbf{x}_t^n, \mathbf{x}_{t-1}, n, y_t)$. A transformer-based denoiser $\mathcal{G}$ is trained to predict the cleaned humanoid state trajectory $x^0_t$ from the noisy input.
\begin{equation}
    \hat{\mathbf{x}}_t^0 = \mathcal{G}(\mathbf{x}_t^n , \mathbf{x}_{t-1}, n, y_t),
\end{equation}
where $\hat{x}_t^0$ is the model’s estimate of the true states. Notably, we predict the states themselves rather than the noise terms, which can improve stability \cite{tevet2024closd}.
During training, terrain embedding is randomly masked by a rate of $0.1$ for classifier-free guidance sampling at inference \cite{ho2021classifier}. 

\textbf{Training.}
We train on our expert locomotion dataset by minimizing the mean-squared error between the predicted and true states. The diffusion reconstruction loss is:
\begin{equation}
    \mathcal{L}_{\text{diffusion}} = \mathbb{E}_{\mathbf{x}_t^0 \sim q(\mathbf{x}_t^0), n \sim [1, N]}[\| \mathbf{x}^0_t - \hat{\mathbf{x}}_t^0\|_2^2].
\end{equation}
During training, to improve stability and performance, we use a scheduled sampling strategy \citep{bengio2015scheduled, yuan2023learning}, where the state history is sampled at the probability $p$ from the ground truth or from the model’s own previous predictions. We gradually anneal the probability $p$ of using true history from 1 to 0 (from fully teacher-forced to fully autoregressive). This curriculum aligns training and inference behavior, preventing divergence in long sequences.

\textbf{Inference.}
After training a denoise network $\mathcal{G}$, we can auto-regressively generate motion imagery given history states $\mathbf{x}_{t-1}$, a standard sampler $\mathcal{S}$ like \citep{ho2020denoising, song2020denoising} and online terrain embeddings $y_t$ by sampling with classifier-free guidance \cite{ho2021classifier}:

\begin{equation}
\begin{split}
    \mathcal{G}_\omega(\mathbf{x}_t^n, \mathbf{x}_{t-1}, n, y_t) &= \mathcal{G}(\mathbf{x}_t^n, \mathbf{x}_{t-1}, n, \phi) \\
    \quad + \omega \cdot \Big( \mathcal{G}(&\mathbf{x}_t^n, \mathbf{x}_{t-1}, n, y_t) - \mathcal{G}(\mathbf{x}_t^n, \mathbf{x}_{t-1}, n, \phi) \Big),
\end{split}
\end{equation}

where $\omega$ is the guidance hyper-parameter, detailed in Appendix~\ref{sup:diffusion}.

\subsection{Diffusion-guided Unified Policy Learning} \label{sec:gcrl}

We formulate the unified policy learning as a Goal-Conditioned Reinforcement Learning (GCRL) problem \cite{liu2022goal}. The goal $g_t$ is the dynamic future trajectory $\hat{s}_{t+1:t+H}$ synthesized by the diffusion world model.
To fully leverage the diffusion world model's predictive power, we design a hybrid reward mechanism that combines precise kinematic tracking with motion manifold regularization.

\textbf{Reward Design.}
Solely tracking the immediate next frame of the diffusion prediction underutilizes the model's ability to forecast coherent long-horizon motions. Conversely, relying solely on style-based rewards can lead to ambiguous optimization objectives. Therefore, we propose a composite reward function:
\begin{equation} \label{eqa:unified_total_reward}
    r_t = w_\text{regu} r_t^\text{regu} + w_\text{track} r_t^\text{track} + w_\text{style} r_t^\text{style},
\end{equation}
where $w_\text{regu}, w_\text{track}, w_\text{style}$ are weighting coefficients.

\textbf{Regularization Reward ($r_t^\text{regu}$):} 
This term enforces fundamental physical constraints, including penalties for excessive joint positions, velocities, accelerations, and torques, as well as encouraging smooth base motion and contact stability.

\textbf{Tracking Reward ($r_t^\text{track}$):} 
Following the unified task-space formulation in BeyondMimic \cite{liao2025beyondmimic}, we define an immediate body-tracking reward using the first predicted step $\hat{s}_{t+1}$ from the diffusion model:
\begin{equation}
r_t^\text{track}
=
\sum_{s \in \{\mathbf{p}, R, \mathbf{v}, \boldsymbol{\omega}\}}
r(\bar{e}_s, \sigma_s),
\qquad
r(\bar{e}_s, \sigma_s)=\exp\left(-\bar{e}_s/\sigma_s^2\right).
\end{equation}
where $\mathcal{B}_{\mathrm{target}}$ denotes the set of tracked bodies, which are represented in a body-relative frame.

\textbf{Style Reward ($r_t^\text{style}$):} The style reward \(r_t^\text{style}\) is computed using an AMP \cite{peng2021amp} approach adapted for online diffusion generation (Algorithm \ref{alg:supp_amp_diffusion}).
A discriminator \(D\) is jointly trained to distinguish between the policy's actual transition \((s_t, s_{t+1})\) and the "dreamed" transition \((\hat{s}_t, \hat{s}_{t+1})\) predicted by the diffusion model. The objective of our discriminator is formulated as:
\begin{equation} \label{eqa:reward_discriminator}
\begin{split}
    \arg\min_{D} \mathbb{E}_{(\hat{s}_t, \hat{s}_{t+1}) \sim p_\text{diff}} \left[ (D (\hat{s}_t, \hat{s}_{t+1})-1)^2\right] + \mathbb{E}_{(s_t, s_{t+1}) \sim\pi}\left[(D(s_t, s_{t+1})+1)^2\right],
\end{split}
\end{equation}

where \(\hat{s}_t\), \(\hat{s}_{t+1}\) are state transitions sampled from diffusion model, \(s_t, s_{t+1}\) are the policy rollouts.
The style reward is then defined as:
\begin{equation}\label{eqa:amp-reward-main}
r_t^\text{style} = \max
\left\{ 0, 1- 0.25(D(\Phi(s_t),\Phi(s_{t+1}))-1)^2 \right\}, 
\end{equation}
which encourages the policy to produce trajectories that align with the diffusion model's implicit knowledge of stable, adaptive locomotion over the prediction horizon. The standard AMP \cite{peng2021amp} relies on offline-sampled action sequences from fixed datasets; our method uniquely employs online inference conditioned on the robot's real-time state \(s_{t-H+1:t}\) to generate the sequence set \(\hat{s}_{t+1:t+K}\). Therefore, a more powerful discriminator is trained on real-time policy trajectories and diffusion-filtered style samples. 

Overall, excluding the shared embodiment-level regularization and safety terms, the unified policy introduces only \textbf{5 non-shared reward terms}: four diffusion-guided body-space tracking terms and one online diffusion-AMP style term. In contrast, TTT-Parkour~\cite{zhu2026ttt} relies on \textbf{14 non-shared reward terms}, including six task-level locomotion rewards, seven terrain/posture-specific shaping rewards, and one AMP-style reward. This comparison shows that our method substantially reduces task-specific reward engineering by replacing handcrafted terrain-dependent terms with diffusion-guided future tracking and online motion-manifold regularization. More details are provided in Appendix~\ref{sup:unified}.

\section{Experiment} \label{sec:experiment}

\begin{figure*}[t]
    \centering
    \includegraphics[width=0.51\linewidth]{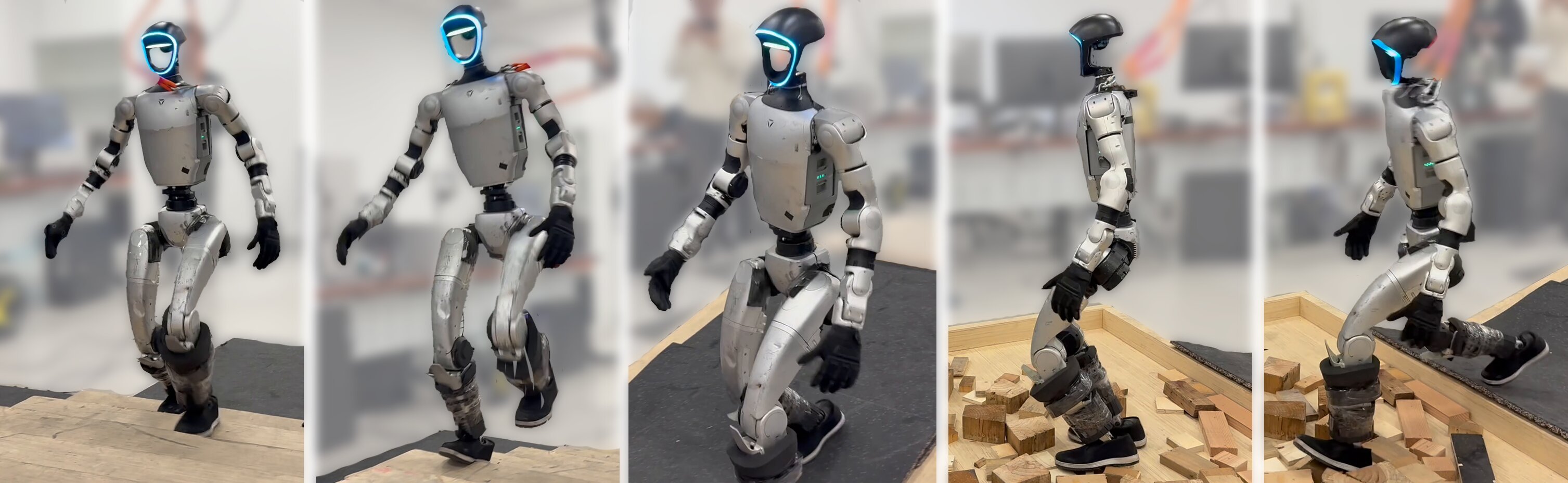}\hfill
    \includegraphics[width=0.48\linewidth]{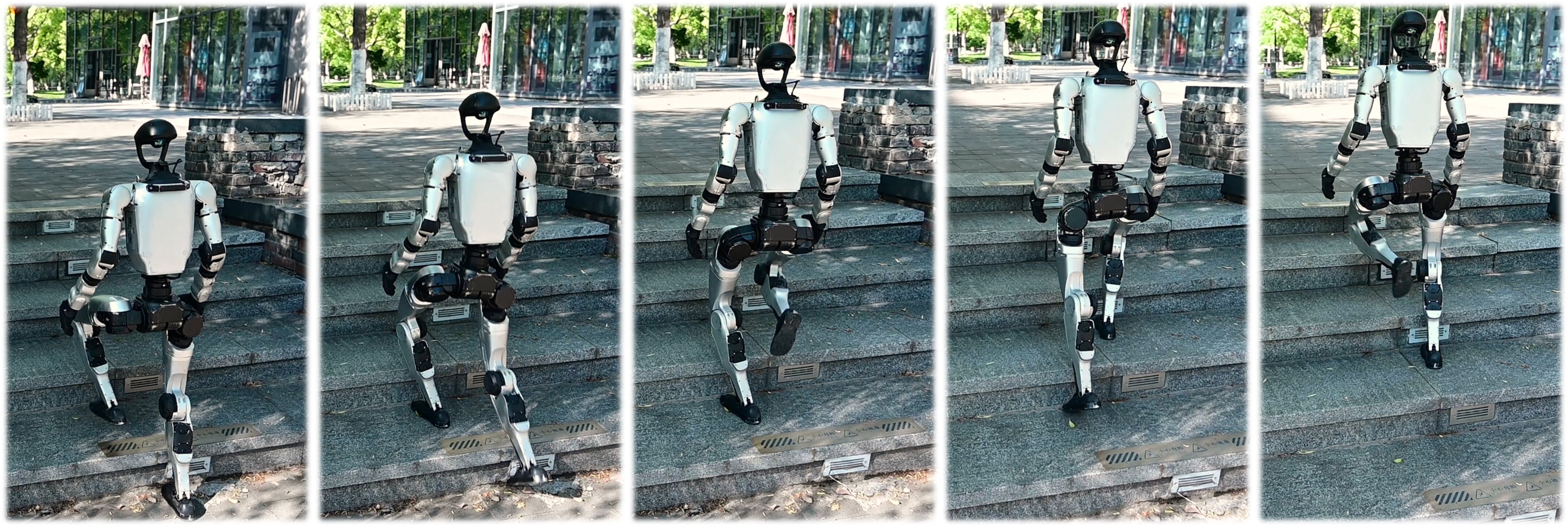}
    \caption{\textbf{Real-world experiment.} We test our policy in the real-world terrains (including stairs, platforms, slopes, and uneven terrain), similar to the terrain in the simulation.}
    \label{fig:real_demo_main}
\end{figure*}

In this section, we conduct comprehensive experiments to evaluate DreamPolicy across various complex terrain scenarios. We begin by detailing the setup of our new humanoid locomotion benchmarks with different terrains. Next, we compared our method with the previous works to demonstrate its superiority. The ablation studies show the contribution of individual components and a scaling analysis to investigate how DreamPolicy benefits from increased data. Finally, we deploy DreamPolicy to real-world scenarios to evaluate its sim2real performance.

\subsection{Simulation Experiments}

\subsubsection{Experimental Setup}

\begin{table*}[t]
\centering
\caption{\textbf{Comprehensive results across unseen terrain types.} This table compares the DreamPolicy against four baselines across four unseen terrain types with Success Rate and Complete Rate.}
\label{tab:ablation_result}
\setlength{\tabcolsep}{2.8pt}
\setlength{\heavyrulewidth}{0.05em}
\setlength{\lightrulewidth}{0.05em}
\setlength{\cmidrulewidth}{0.03em}
\resizebox{\textwidth}{!}{
\begin{tabular}{l cccc c cccc}
\toprule
& \multicolumn{4}{c}{\textbf{Success Rate (\%) $\uparrow$}} & & \multicolumn{4}{c}{\textbf{Complete Rate (\%) $\uparrow$}} \\
\cmidrule(lr){2-5} \cmidrule(lr){7-10}
\textbf{Method} & \textbf{Slope bridge} & \textbf{Balancing Beam} & \textbf{Wave} & \textbf{Slope} & & \textbf{Slope bridge} & \textbf{Balancing Beam} & \textbf{Wave}  & \textbf{Slope} \\
\midrule
Ours     & \textbf{92.77} & \textbf{96.35} & \textbf{100.0} & \textbf{100.0} & & \textbf{60.48} & \textbf{90.32} & \textbf{100.0} & \textbf{100.0} \\
Distillation\citep{ross2011reduction} & 65.48 & 74.36 & \textbf{100.0} & \textbf{100.0} & & 40.16 & 60.24 & \textbf{100.0} & \textbf{100.0} \\
w/o diffusion & 0.0  & 20.14 & 97.32 & 98.65 & & 10.35 & 35.78 & 98.46 & 99.16 \\
w/o GCRL   & 0.0  & 0.0  & 60.38 & 94.35 & & 1.35  & 5.34  & 80.26 & 90.63 \\

\bottomrule
\addlinespace
\end{tabular}}
\end{table*}

\setlength{\tabcolsep}{4pt}
\begin{table*}[t]
\vspace{-5pt}
\caption{\textbf{Performance Comparison in combined terrain.} All data in the table are shown as percentages, the higher the better. The best results are highlighted in boldface.}\label{add:multi_2_2}
\begin{center}
\resizebox{\linewidth}{!}{
\begin{tabular}{l cccccccccc}
\toprule[1.0pt]
\multicolumn{1}{l}{\multirow{2}{*}{}}  
& \multicolumn{2}{c}{Stair\&Uneven} 
& \multicolumn{2}{c}{Bridge\&Uneven}  
&\multicolumn{2}{c}{Stair\&Gap}  
&\multicolumn{2}{c}{Stair\&Bridge} 
&\multicolumn{2}{c}{Bridge\&Gap}\\
\cmidrule[\heavyrulewidth](lr){2-3} \cmidrule[\heavyrulewidth](lr){4-5} \cmidrule[\heavyrulewidth](lr){6-7} \cmidrule[\heavyrulewidth](lr){8-9} \cmidrule[\heavyrulewidth](lr){10-11}
\multicolumn{1}{l}{} 
& $R_{\mathrm{succ}}(\%)$ & $R_\mathrm{cmp}(\%)$ 
& $R_{\mathrm{succ}}(\%)$ & $R_\mathrm{cmp}(\%)$ 
& $R_{\mathrm{succ}}(\%)$ & $R_\mathrm{cmp}(\%)$  
& $R_{\mathrm{succ}}(\%)$ & $R_\mathrm{cmp}(\%)$ 
& $R_{\mathrm{succ}}(\%)$ & $R_\mathrm{cmp}(\%)$\\
\midrule[0.8pt]

Distillation\citep{ross2011reduction} 
& $50.48$ & $76.45$ 
& $62.33$ & $74.58$ 
& $88.42$ & $90.35$ 
& $86.76$ & $72.17$ 
& $91.34$ & $82.65$\\ 

Ours   
& \textbf{88.46} & \textbf{96.25} 
& \textbf{84.78} & \textbf{90.64} 
& \textbf{95.79} & \textbf{98.68} 
& \textbf{91.45} & \textbf{84.65} 
& \textbf{94.68} & \textbf{96.58}\\ 

\bottomrule[1.0pt]
\end{tabular}
\vspace{-5pt}
}
\end{center}
\end{table*}

\textbf{Comparison Methods.}
We compare our method against the following settings: 1) scenario specialized policy that extends the Actor-Critic architecture from \cite{rudin2022learning} with a scandot encoder. 2) distillation-based method following DAgger \citep{chen2020learning, ross2011reduction}; 3) ablation without the Diffusion Planner; 4) ablation without Goal-Conditioned RL. More design details are provided in the Appendix~\ref{sup:baseline}. 

\textbf{Scenarios.}

We evaluated our method on 1) \textbf{Five single-terrains} \textit{ (Flat, Gap, Stair, Bridge, Uneven)}. 2) \textbf{Four unseen terrains} \textit{(Wave, Slope , Balancing Beam, Slope bridge)}. As shown in Figure \ref{exp:terrain_figure}. More design details are provided in Appendix~\ref{sup:benchmark}.

\textbf{Metrics.}
We evaluated all settings using the following metrics.
1) Success Rate $R_{\text{succ}}$: The percentage of successful attempts to cross the entire terrain. 2) Complete Rate $R_{\text{cmp}}$: The ratio of the distance traveled before falling to the total terrain length. 

\subsubsection{Result} \label{exp:generalization}

\textbf{Single-terrain.} DreamPolicy matches expert performance without the regression issues of distillation baselines (Fig.~\ref{exp:single_terrain_figure}, Table~\ref{add:single_terrain}), effectively learning a unified locomotion policy.
\textbf{Unseen terrains.} For zero-shot generalization on four novel terrains (Table~\ref{tab:ablation_result}), our method demonstrates superior performance over prior approaches.
\textbf{Combined terrains.} Under compounding complexity (Table~\ref{add:multi_2_2}), DreamPolicy significantly outperforms baselines. The gap is notably striking on challenging combinations: yielding 37.98\% and 22.45\% improvements over the baseline on 'Stair\&Uneven' and 'Bridge\&Uneven', while maintaining >91\% success elsewhere.
Overall, DreamPolicy exhibits strong generalization and robustness in complex environments (Appendix~\ref{sup:visual}).

\subsubsection{Ablation Analysis}
To isolate the contribution of each individual component in DreamPolicy, we conduct ablation studies on Diffusion World Model and the Goal-Conditioned RL (GCRL) module, respectively.

\textbf{Ours w./w.o. Diffusion World Model.}
We first evaluate DreamPolicy without the diffusion planner, forcing the agent to rely exclusively on the aggregated offline expert data. As shown in Table \ref{tab:ablation_result}, this leads to a sharp decline in performance, which worsens rapidly as terrain complexity increases. Without the planner's guidance, the policy fails to generalize to out-of-distribution (OOD) states. Without the planner to provide future trajectories, the method cannot adapt to unseen complex environments.

\textbf{Ours w./w.o. GCRL.}
Ablation results without GCRL indicate that relying solely on the diffusion model performs well on simple terrains (Figure \ref{exp:single_terrain_figure}) but fails on more complex environments (Table \ref{tab:ablation_result}). While the diffusion model can generate optimal policies, these are not always physically feasible and can be disrupted by environmental factors, such as stumbling on uneven ground. Notably, BiRoDiff \cite{mothish2024birodiff}, which uses a similar architecture, exhibits consistent results, further supporting our findings.

\subsubsection{Scaling Analysis}

\begin{figure}[t]
  \centering
  \begin{subfigure}[t]{0.52\textwidth}
    \centering
    \includegraphics[width=\linewidth]{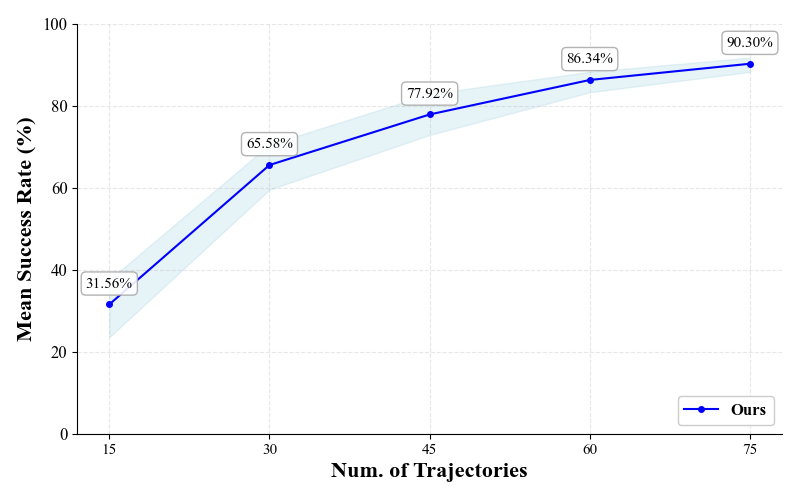}
    \caption{Scaling law of numbers}
    \label{exp:scaling_nums}
  \end{subfigure}\hfill
  \begin{subfigure}[t]{0.44\textwidth}
    \centering
    \includegraphics[width=\linewidth]{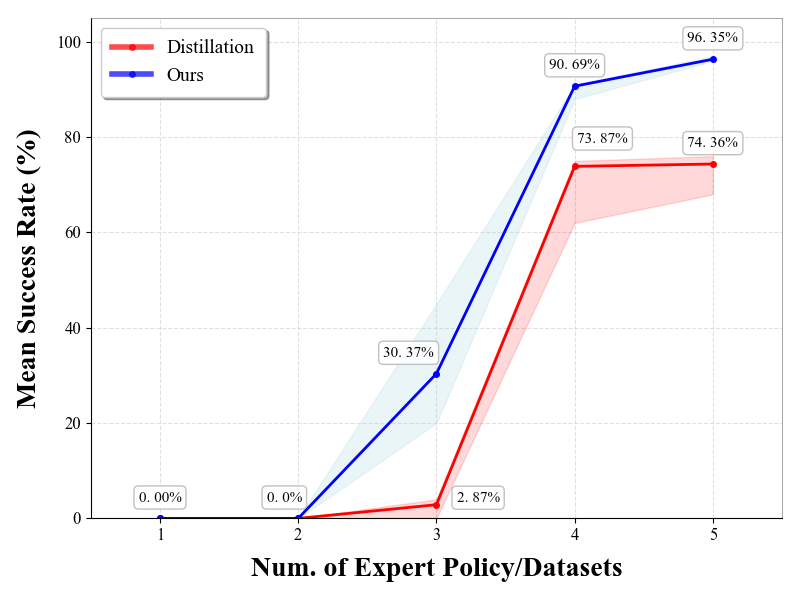}
    \caption{Scaling law of types}
    \label{exp:scaling_type}
  \end{subfigure}
  \caption{\textbf{Scalability Analysis for dataset size and types.} Both subfigures demonstrate a consistent increase in the mean success rate as the offline dataset scales up in either (a) size or (b) diversity.}
  \label{exp:scaling_overall}
\end{figure}

We evaluate DreamPolicy's scalability on the unseen terrain. Fig.~\ref{exp:scaling_nums} shows consistent performance gains as the dataset size scales from 5M to 75M transitions. Furthermore, Fig.~\ref{exp:scaling_type} evaluates scalability across dataset diversity (1 to 5 terrain types). DreamPolicy exhibits significantly higher data efficiency than standard distillation. 

\subsection{Real-world Experiments}

\subsubsection{Experimental Setup} 
We deploy the proposed framework on the Unitree G1 humanoid robot using a distributed ROS architecture to ensure strict Sim-to-Real alignment. The system utilizes a distributed setup: the robot's onboard computer handles raw perception (FAST-LIO2 \cite{xu2022fast}), while an external PC executes the computationally intensive model inference. To match the sparse scandots ($o_t^\text{terrain}$) used during training, raw LiDAR data is processed via elevation mapping \cite{fankhauser2018probabilistic}, downsampled, and cropped to a robot-centric height grid ($[-0.6, 1.0]$m longitudinal, $\pm0.5$m lateral). This processed map is published at approximately 20Hz. To mitigate latency, we implement an asynchronous buffering mechanism that decouples the 50Hz control policy from the 20Hz perception and $\approx$20ms inference latency. By strictly fetching the latest available observations from the buffer, the system maintains robust stability without blocking.

\subsubsection{Qualitative results}
We evaluate the policy across diverse real-world terrains, including stairs, slopes, gaps, uneven surfaces, and their complex combinations. We highly recommend watching the supplementary video for dynamic demonstrations of these experiments.

\textbf{Zero-shot Sim-to-Real Transfer.}
As illustrated in Fig.~\ref{fig:real_demo_main}, the robot successfully performs robust locomotion across all tested scenarios via zero-shot transfer. Despite the discrepancy between the simulation physics and the real-world dynamics of the Unitree G1, the proposed framework demonstrates strong generalizability. Notably, the robot is able to traverse discrete gaps and climb stairs continuously without pausing, effectively managing the transition between flat ground and uneven terrain.

\textbf{Adaptability to Composite Terrains.}
A key challenge in our experiments is the transition between different terrain types (e.g., from a flat platform to a stair, or across mixed obstacles). As shown in Fig.~\ref{fig:composite_terrain}, the diffusion world model plays a critical role here. By synthesizing terrain-adaptive reference trajectories, the system anticipates future kinematic states, allowing the policy to adjust the robot's torso posture and foot placement. This predictive capability is crucial for maintaining balance on composite terrains.

\section{Conclusion, Limitations and Future Works} \label{sec:limitation}
We propose DreamPolicy, a unified framework that enables versatile humanoid locomotion across diverse scenarios and unseen terrains. Our approach begins by collecting expert data via parallel simulations. Subsequently, a terrain-aware autoregressive diffusion world model is trained on this dataset to synthesize terrain-adaptive reference trajectories, effectively capturing rich, multimodal motion priors. Finally, a physics-constrained RL policy is optimized using this diffusion-generated guidance, ensuring both dynamical feasibility and robust exploration. 
By combining scalable skill acquisition with probabilistic motion modeling, DreamPolicy achieves broader state–action coverage and superior generalization. However, a primary limitation lies in the computational overhead of the diffusion world model, which currently relies on iterative denoising. Future work will explore lightweight sampling techniques to accelerate inference speeds, enabling synchronous, high-frequency planning without sacrificing trajectory quality.

\section*{Acknowledgement}
This work was supported by the National Natural Science Foundation of China (62303319, 62406195), Shanghai Local College Capacity Building Program (23010503100), ShanghaiTech AI4S Initiative SHTAI4S202404, HPC Platform of ShanghaiTech University, and MoE Key Laboratory of Intelligent Perception and Human-Machine Collaboration (ShanghaiTech University), and Shanghai Engineering Research Center of Intelligent Vision and Imaging. This work was also supported in part by computational resources provided by Fcloud CO., LTD.

\newpage

\bibliographystyle{plainnat}
\bibliography{references}

\newpage
\clearpage

\appendix

\onecolumn
\begin{center}
{ \linespread{1.5} \selectfont
\textbf{\Large DreamPolicy: A Unified World-model Policy for Scalable Humanoid Locomotion} \\
}
\linespread{1.5} \Large Supplementary Material
\end{center}
\setcounter{table}{0}  
\setcounter{figure}{0}

\renewcommand{\thetable}{A\arabic{table}}
\renewcommand{\thefigure}{A\arabic{figure}}
\makeatletter

\startcontents[appendix]
\printcontents[appendix]{l}{1}{\section*{Appendix}\setcounter{tocdepth}{2}}

\section{Overview}
\label{sup:overview}

The appendix provides comprehensive supplementary materials supporting the methodological and empirical contributions of this work. Section~\ref{sup:technical} elaborates on technical implementations, including the configuration of humanoid locomotion benchmarks (\ref{sup:benchmark}), specialized policy training protocols (\ref{sup:specialist}), architectural details of the autoregressive diffusion planner (\ref{sup:diffusion}), and the design principles of the diffusion guided unified policy (\ref{sup:unified}). Section~\ref{sup:result} presents extended experimental analyses, encompassing baseline implementation specifics (\ref{sup:baseline}), additional performance evaluations across diverse scenarios (\ref{sup:experiments}), qualitative visualizations (\ref{sup:visual}), and supplementary video demonstrations (\ref{sup:video}). Section~\ref{sup:conclusion} concludes with an in-depth discussion of limitations through failure case examinations (\ref{sup:limitation}) and potential future research directions (\ref{sup:discussion}).

\section{Supplementary Technical Details}
\label{sup:technical}

\subsection{Details of Humanoid Locomotion Benchmarks}
\label{sup:benchmark}
\paragraph{Curriculum design}

For the parameters in the table, $p[0]$ represents the first dimension, $p[1]$ represents the second dimension. If $p[0] > p[1]$, the larger the parameter, the simpler it is. The curriculum design is as follows:
\begin{equation}
\begin{split}
\text{Truth parameters} = \text{random} \biggl(  \frac{p[1]-p[0]}{\text{max\_terrain\_level}} \times \text{Difficulty} \biggr) + p[0]
\end{split}
\end{equation}

\paragraph{Detailed parameters of different terrains.}
For all terrains, we add base roughness to make the ground undulating. The height of the ground is between $\text{random}(-0.02,0.02)\times \text{Difficulty}$. The difficulty increases with terrain level.
Notably, prior research typically used test terrains longer than 4m \cite{xie2025humanoid} and 8m \cite{wang2025beamdojo} for single terrains. To ensure a more robust evaluation and minimize random effects, we increase the test terrain length to 10m. 

The five Single-Terrains' specific parameters are shown in Table \ref{bench:single_p}. For the \textit{Gap}, the Size Range represents the platform that robots can stand, the gap robots need to cross is $[0.2,0.3]$. 

\begin{table*}[ht]
\centering
\caption{\textbf{Single-Terrain curriculum parameter.} We designed different curriculum parameters for different Single-Terrains.}
\label{bench:single_p} 
\begin{tabular}{@{}ccccc@{}}
\toprule 
& Parameter & \textit{Height\_range (m)} & \textit{Size\_range (m)} & \textit{Subsection\_nums} \\
\midrule 
\multirow{5}{*}{\makecell[c]{Single \\ Terrain}} 
& Flat   & --             & --             & --          \\
& Uneven & $[0.10, 0.20]$ & $[0.40, 0.70]$ & $[150, 200]$ \\
& Stair  & $[0.15, 0.30]$ & $[0.40, 0.25]$ & $[5, 10]$    \\
& Gap    & --             & $[0.60, 0.40]$ & $[5, 10]$    \\
& Bridge & --             & $[0.40, 0.20]$ & --          \\
\bottomrule
\end{tabular}
\end{table*}

Due to the particularity of the four \textbf{Unseen-Terrains}, we will introduce each terrain separately.
\begin{itemize}
    \item The \textit{Wave} adds several sinusoidal functions to make the ground undulate. The number of sine functions $[5,10]$, and the amplitude of each sine function $[0.2,0.5]$. This terrain is designed to test whether the robot can maintain a stable center of gravity when facing disturbances from irregular ground.

    \item The \textit{Slope} had a fixed inclination angle of 30°. The surface was covered with a coarse-grit sandpaper to provide a consistent, high-friction interface with a measured coefficient of static friction.

    \item In \textit{Balancing Beam}, the width of the standing platform across the gap is set to match the bridge. This configuration is designed to evaluate how well the robot integrates two distinct locomotion strategies.

    \item The \textit{Slope-Bridge} is a bridge whose height is set according to the first half of the sine function. The slope of an arch bridge is determined by the length and amplitude of the terrain. Since we have fixed the length of the terrain to 18 meters, the amplitude is set to $[1.5,2.5]$ meters, with the width as a single bridge. This terrain is designed to test the robot's ability to generalize from stair terrain to slopes.

\end{itemize}
\paragraph{Cross embodiment and Cross simulation}

To support the research community's transition toward next-generation simulation standards, we provide distinct, fully validated implementations of our framework for both Isaac Gym and Isaac Lab (NVIDIA Omniverse). While the Isaac Gym version offers a highly optimized backend for rapid, high-throughput training on consumer hardware, the Isaac Lab implementation leverages the advanced PhysX 5 engine and high-fidelity rendering to enhance simulation accuracy and Sim-to-Real transfer. Both implementations adhere to a strictly aligned benchmark with identical observation spaces, reward formulations, and terrain generation to ensure algorithmic training across different simulation platforms.

We demonstrate the cross-embodiment scalability of DreamPolicy by validating it on distinct robots, including the Unitree H1-2 and Unitree G1. The inclusion of specialized configuration presets for these platforms confirms that our approach generalizes effectively across dynamic systems without overfitting to specific kinematics. Furthermore, the modular design of our reward and observation structures allows researchers to seamlessly integrate new URDF/MJCF models, facilitating the rapid benchmarking of novel hardware platforms within our standardized environments.

\begin{figure*}[ht]
    \centering

    \begin{subfigure}[b]{0.48\linewidth}
        \centering
        \includegraphics[width=\linewidth]{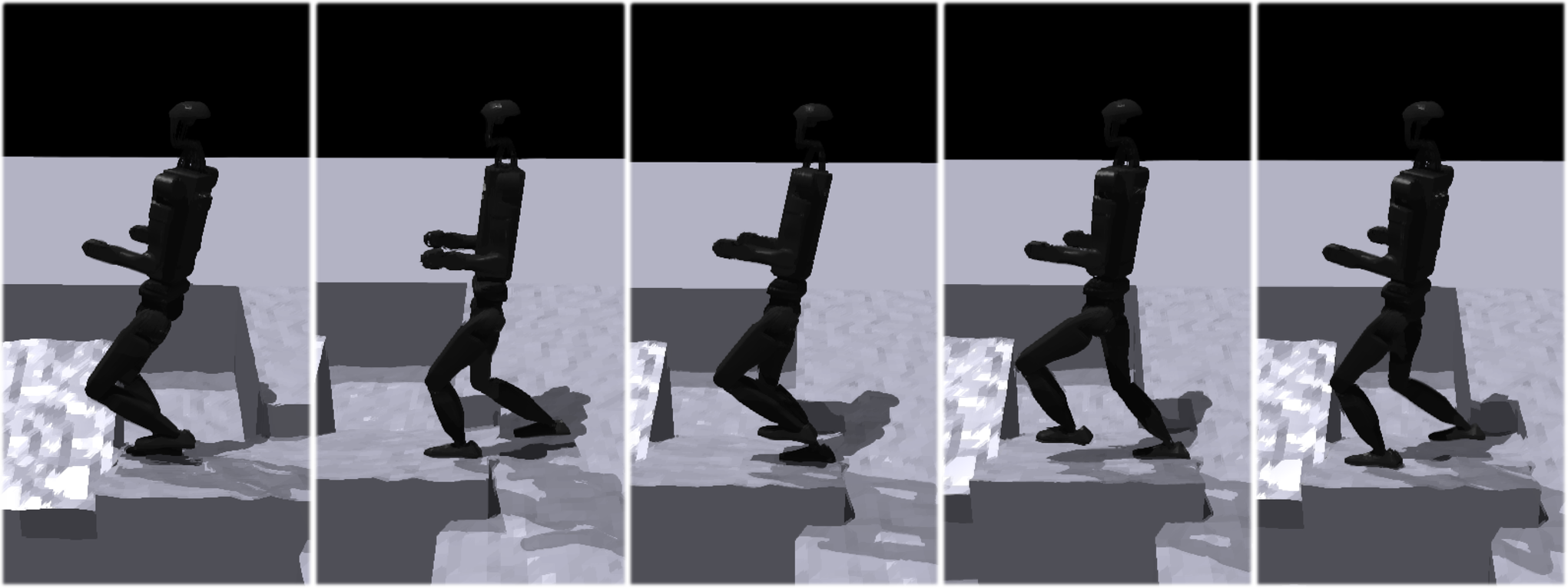}
        \caption{Simulation qualitative result on parkour terrain.}
        \label{fig:single1}
    \end{subfigure}
    \hfill
    \begin{subfigure}[b]{0.48\linewidth}
        \centering
        \includegraphics[width=\linewidth]{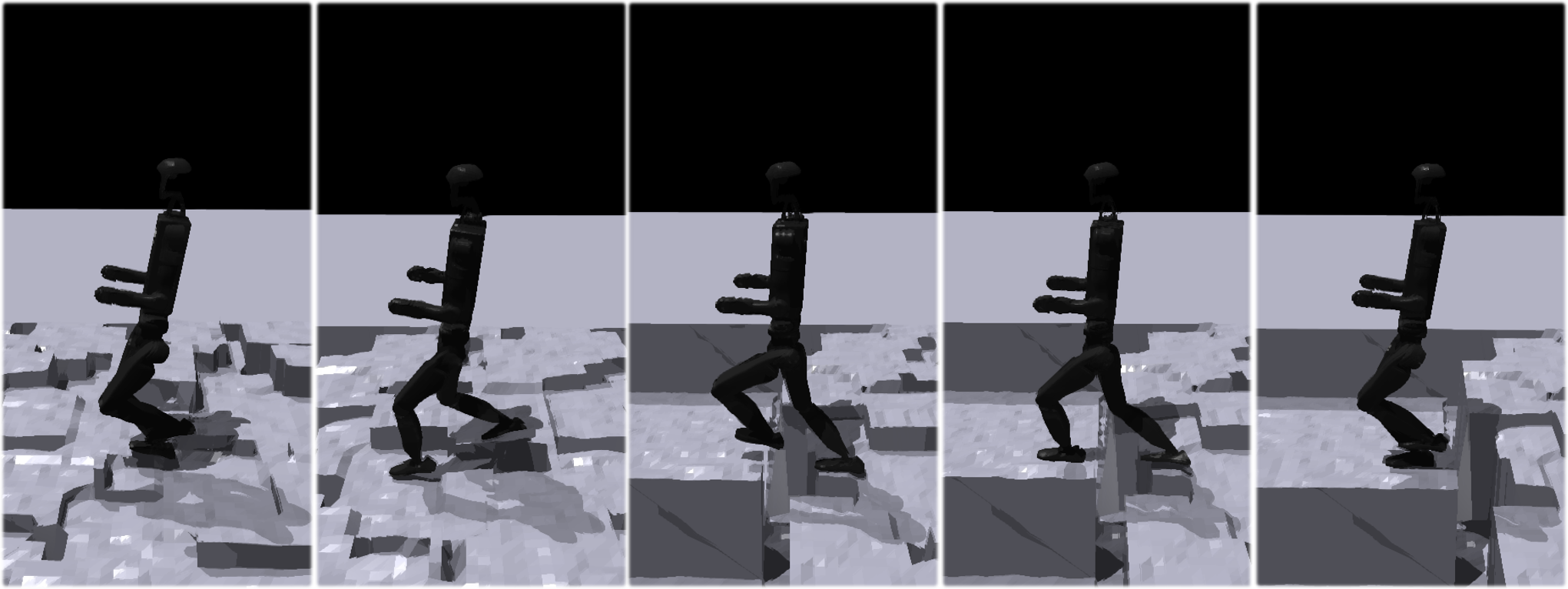}
        \caption{Simulation qualitative result on uneven terrain.}
        \label{fig:single2}
    \end{subfigure}

    \vspace{0.5em}

    \begin{subfigure}[b]{0.48\linewidth}
        \centering
        \includegraphics[width=\linewidth]{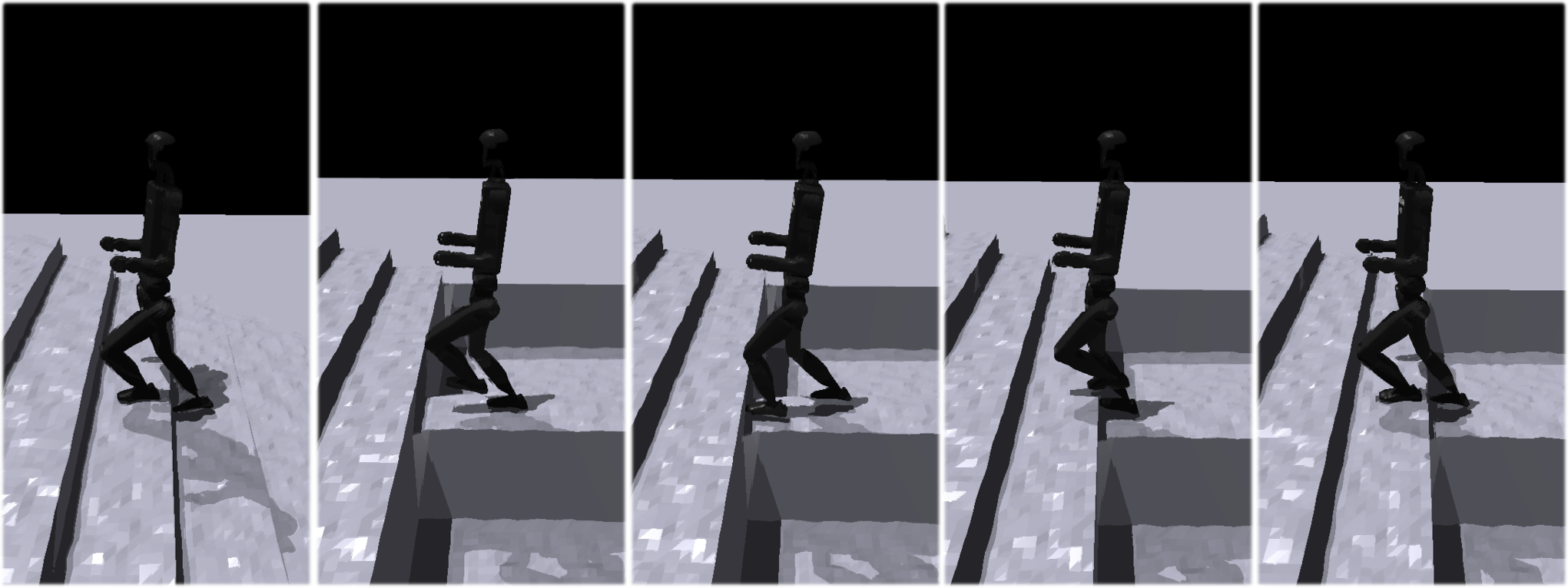}
        \caption{Simulation qualitative result on bridge terrain.}
        \label{fig:single3}
    \end{subfigure}
    \hfill
    \begin{subfigure}[b]{0.48\linewidth}
        \centering
        \includegraphics[width=\linewidth]{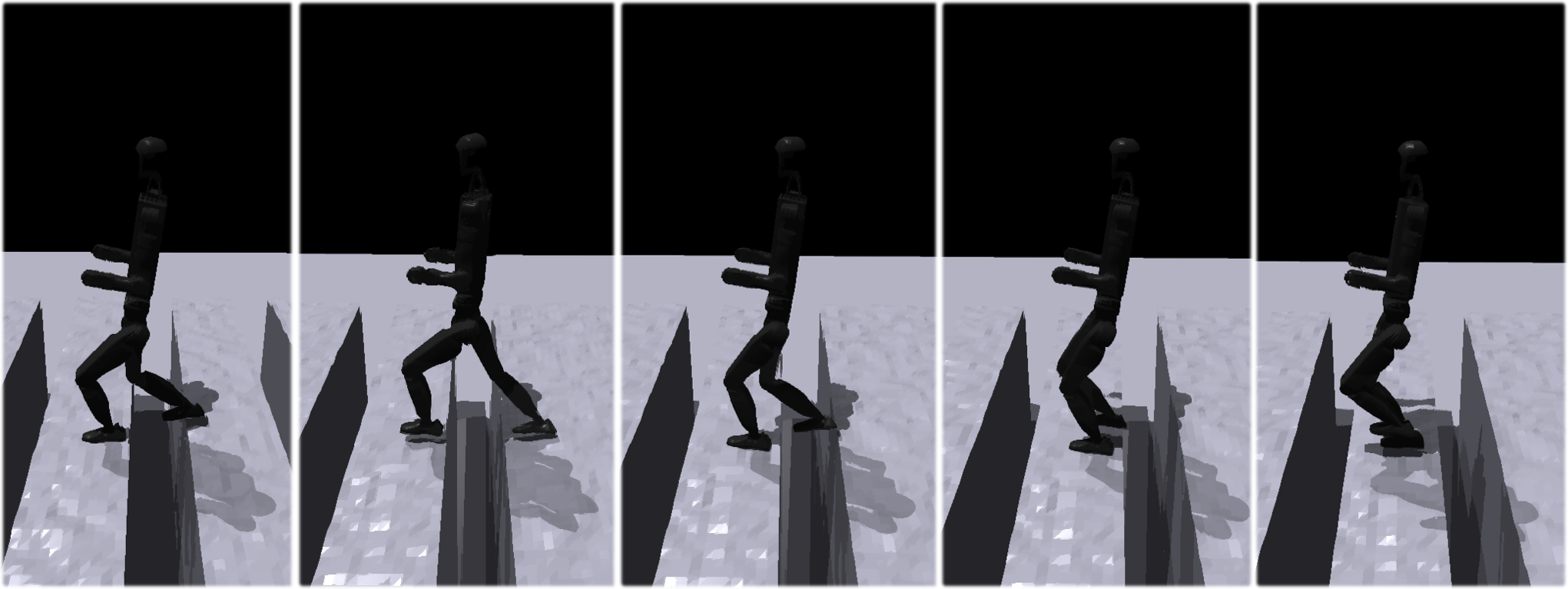}
        \caption{Simulation qualitative result on gap.}
        \label{fig:single4}
    \end{subfigure}
    \caption{Simulation qualitative result in IsaacGym.}
    \label{fig:sim-result-appendix2}
\end{figure*}

More visualization results are shown in Fig.~\ref{fig:sim-result-appendix2}, and Figure \ref{fig:sim-result-appendix1}.

\paragraph{Detailed expert data collection.}
We collected expert data from three episodes in five terrains, six rounds each, and 1024 environments, totaling about 75M frames of data. In each frame, we record the following information, as shown in the Table \ref{bench:dataset}. 

\begin{table}[ht]
    \centering
    \caption{Robot Locomotion Expert Data Items.}
    \begin{tabular}{c|c}
    \hline
        Terrain Scandot & (T, 187) \\
        Robot Observations & (T, 98) \\
        Actions & (T, 29) \\
    \hline
    \end{tabular}
    \label{bench:dataset}
\end{table}

\subsection{Details of Specialized Policy Learning}\label{sup:specialist}

\paragraph{Model Architecture}

For each specialized ideal policy, we formulate it using an MLP as the backbone \(\mathcal{F}\) of the actor. Perceptive scandots is encoded by an MLP encoder \(\mathcal{E}_\text{terrain}\). Apart from that, an RNN-based encoder \(\mathcal{E}_\text{hist}\) is applied to construct the latent information of the history state sequence. Details of model architecture are listed in Table \ref{tab:supp-special-archi}.

The actor is formulated as:
\begin{equation} \label{eqa:policy_archi}
    a_t = \mathcal{F} \left( \mathcal{E}_\text{hist}(o_t^{\text{hist}}), \mathcal{E}_\text{scan}(o_t^\text{terrain}), o_t^\text{proprio}, c_t, a_{t-1} \right).
\end{equation}
The critic is also formulated as a MLP \(\mathcal{C}\):
\begin{equation}
v_t=\mathcal{C}\left(\mathcal{E}_\text{hist}(o_t^\text{hist}),\mathcal{E}_\text{scan}(o_t^\text{terrain}),o_t^\text{proprio},c_t,a_{t-1}\right).
\end{equation}
The reinforcement learning algorithm used to train the Goal-conditioned policy is Proximal Policy Optimization Algorithm \cite{ppo}. The hyperparameters of the PPO Algorithm are listed in Table \ref{tab:supp-ppo-param} of the supplementary. We use the same hyperparameters for the training of all specialized policies.

\begin{table}[ht]
    \centering
    \caption{Specialized Policy Structure}
    \label{tab:supp-special-archi}
    \begin{tabular}{cc}
    \toprule
    Parameter & Value \\
    \midrule
    Actor Hidden Dims & [512, 256, 128] \\
    Critic Hidden Dims & [512, 256, 128] \\
    Scan Encoder Dims & [128, 64, 32] \\

    RNN Hidden Size & 512 \\
    RNN Layers & 1 \\
    \bottomrule
    \end{tabular}
\end{table}

\begin{table}[ht]
    \centering
    \caption{PPO Hyperparameters for Specialized Policy Training}
    \begin{tabular}{cc}
    \toprule
        Hyperparameter & Value \\
    \midrule
        Clip range & 0.2 \\
        GAE discount factor & 0.95 \\
        Entropy coefficient & 0.01 \\
        Max gradient norm & 1.0 \\
        Learning rate & 2e-4 \\
        Reward discount factor & 0.99 \\
        Initial policy std & 1.0 \\
        Minimum policy std & 0.2 \\
        Number of environments & 2048 \\
        Number of environment steps per training batch & 24 \\
        Learning epochs per training batch & 5 \\
        Number of mini-batches per training batch & 4 \\
    \bottomrule
    \end{tabular}
    \label{tab:supp-ppo-param}
\end{table}

\paragraph{Reward Functions for Specialized Policy}

A walk policy is first trained from scratch through reinforcement learning. Subsequently, a series of terrain-specific policies is developed for individual challenging terrains. To enable this progressive learning capability, we design two distinct reward categories. Basic locomotion rewards, as illustrated in Table \ref{tab:supp_reward_base}, which enforces fundamental kinematic constraints and penalizes excessive joint torque/velocity for feasible actuator commands. Apart from that, we design specialized rewards for challenging terrains, as shown in Table \ref{tab:supp_reward_special}, to provide harder constraints on various scenarios.

\begin{table*}[t]
    \centering
    \caption{\textbf{Basic Reward Functions and Weight } (Notations: $\boldsymbol{v}$=velocity vector, $\omega$=angular velocity, $yaw$=heading angle, $F$=contact force, $\theta$=joint angle, $\tau$=torque, $h$=height, $a$=action, $\mathbb{I}(\cdot)$=indicator function)
    }
    \begin{tabular}{lcl}
    \toprule
        \textbf{Reward Function} & \textbf{Formula} & \textbf{Weight} \\
    \midrule

        Orientation & 
        $\sum (g_x^2 + g_y^2) $ & 
        -1.0 \\
        DOF Acceleration & 
        $\sum \left(\frac{\dot{\theta}_{t} - \dot{\theta}_{t-1}}{\Delta t}\right)^2$ & 
        -2.5e-7 \\
        Collision & 
        $\sum \mathbb{I}(\|F_{\text{contact}}\| > 0.1)$ & 
        -10.0 \\
        Action Rate & 
        $\|a_t - a_{t-1}\|$ & 
        -0.1 \\
        Delta Torques & 
        $\sum (\tau_t - \tau_{t-1})^2$ & 
        -1.0e-7 \\
        Torques & 
        $\sum \tau^2$ & 
        -0.00001 \\
        DOF Error & 
        $\sum (\theta - \theta_{\text{default}})^2$ & 
        -0.04 \\

        Base Height & 
        $(h_{\text{base}} - h_{\text{target}})^2$ & 
        -20.0 \\

        Smoothness & 
        $\|a_t - a_{t-1}\| + \|(a_t - a_{t-1}) - (a_{t-1} - a_{t-2})\|$ & 
        -0.005 \\
        Alive Bonus & 
        $\sum(\text{survival})$ & 
        0.005 \\
        Tracking Yaw & 
        $\exp(-|yaw_{\text{target}} - yaw|^2 / 0.25)$ & 
        0.5 \\
        Linear Velocity Z & 
        $v_z^2 $ & 
        -1.0 \\
        Angular Velocity XY & 
        $\sum ({\omega_x^2 + \omega_y^2})$ & 
        -0.05 \\
    \bottomrule
    \end{tabular}
    \label{tab:supp_reward_base}
\end{table*}

\begin{table*}[ht]
    \centering
    \caption{\textbf{Special Reward Functions and Weight}.}
    \begin{tabular}{lcl}
    \toprule
        \textbf{Reward Function} & \textbf{Formula} & \textbf{Weight} \\
    \midrule
       Tracking Goal Velocity & 
        $\min\left(\frac{\boldsymbol{v}_{\text{cur}} \cdot \boldsymbol{v}_{\text{target}}}{\|\boldsymbol{v}_{\text{target}}\|}, v_{\text{cmd}}\right) \big/ v_{\text{cmd}}$ & 
        3.0 \\
        Hip Position & 
        $\sum (\theta_{hip} - \theta_{\text{default}})^2$ & 
        -0.5 \\
         Feet Stumble & 
        $\mathbb{I}(\|F_{\text{horizontal}}\| > 4|F_z|)$ & 
        -1.0 \\
        Feet Edge & 
        $\sum \mathbb{I}(\text{foot\_on\_edge}) \times \mathbb{I}(\text{terrain\_level}>3)$ & 
        -1.0 \\
        Feet Parallel & 
        $\sum \theta_{\text{pitch}}^2$ & 
        -1.0 \\
        Huge Step & 
        $\sum \ abs(x_{\text{left\_ankle}}-x_{\text{right\_ankle}})$ & 
        0.001 \\
        Feet Yaw & 
         $\sum \ ({yaw}_{\text{ankle}})^2 $ & 
        -2. \\
         Feet Dis & 
         $\sum \ (({y}_{\text{left\_ankle}})-({y}_{\text{right\_ankle}}))^2 $ & 
        -0.001 \\
        Knee Dis & 
         $\sum \ (({y}_{\text{left\_knee}})-({y}_{\text{right\_knee}}))^2 $ & 
        -0.001\\
        Goal & 
         $\sum \ (goal) $ & 
        0.0002 \\
        Knee Foot & 
         $\sum \ (({y}_{\text{knee}})-({y}_{\text{ankle}}))^2 $ & 
        -0.001 \\
        Foot Height& 
         $\sum \ ({z}_{\text{ankle}}) $ & 
        0.01 \\
        Foot Air Time& 
         $\sum \ ({t}_{\text{foot\_contact==0}}) $ & 
        10 \\
    \bottomrule
    \end{tabular}
    \label{tab:supp_reward_special}
\end{table*}

\subsection{Details of Autoregressive Diffusion World Model}
\label{sup:diffusion}

\paragraph{Terrain-Aware Autoregressive Diffusion Training}
To bridge the gap between training and autoregressive inference, we employ a three‑stage scheduled training \cite{bengio2015scheduled, yuan2023learning, zhao2024dartcontrol} curriculum. In our terrain‑aware autoregressive diffusion world model, we organize training data into overlapping motion windows of fixed length $H$ and slide them through each sequence. At each training iteration $iter$, we compute a rollout probability $p$ in Alg. \ref{alg:diffusion_rollout}. We set a maximum of 4 consecutive rollouts during each update to stabilize training and prevent error accumulation.

\begin{algorithm*}[tb]
\caption{Compute schedule rollout probability}
\label{alg:diffusion_rollout}
\begin{algorithmic}[1]
    \State {\bfseries Input:} current iteration step $iter$, number of iterations in the fully supervised phase $I_1$, number of iterations in scheduled phase $I_2$
    \State {\bfseries Output:} probability $p$
    \State
    \State {\bfseries Function} \textsc{Schedule\_probability}($iter$, $I_1$, $I_2$)
    \If{$iter \leq I_1$}
        \State $p \gets 0$ \hfill {\color{gray}\textit{// Fully supervised phase}}
    \ElsIf{$iter > I_1 + I_2$}
        \State $p \gets 1$ \hfill {\color{gray}\textit{// Rollout training phase}}
    \Else
        \State $p \gets \frac{iter - I_1}{I_2}$ \hfill {\color{gray}\textit{// Scheduled training phase}}
    \EndIf
    \State \textbf{return} $p$
\end{algorithmic}
\end{algorithm*}

In our terrain‑aware autoregressive diffusion model, we adopt a complementary three‑phase annealing: at each iteration, we sample the history from prior model predictions with probability $p$ and from the ground truth with probability $1-p$, gradually increasing $p$ from $0$ to $1$:

\begin{itemize}
    \item \textbf{Fully supervised phase.} $p = 0$. Always use ground‑truth history, which is fully teacher-forced.
    \item \textbf{Scheduled Training Phase.} $p$ linearly increases from $0$ to $1$. With probability $p$, replace the ground‑truth history with the model’s own rollout history; otherwise keep the ground truth.
    \item \textbf{Rollout Training Phase.}  $p = 1$. Always use rollout history, which is fully autoregressive.
\end{itemize}
This simple curriculum likewise exposes the model to its inference‑time distribution early in training, improving long‑horizon stability and preventing sequence drift. The training algorithm is shown in Alg. \ref{alg:diffusion_train}.

\begin{algorithm*}[t]
\caption{Scheduled training for terrain-aware auto-regressive diffusion}
\label{alg:diffusion_train}
\begin{algorithmic}[1]
\State \textbf{Input:} denoiser $\mathcal{G_\theta}$ with parameters $\theta$, humanoid state dataloader $\mathcal{X}$, terrain dataloader $\mathcal{Y}$, total diffusion steps $N$, consecutive window number $T$, optimizer $\mathcal{O}$, training loss $\mathcal{L}$, max iteration $I_{max}$, num of iterations in the fully supervised phase $I_1$, num of iterations in scheduled phase $I_2$, auto-regressive diffusion sampler $\mathcal{S}$.
\State $iter \gets 0$
\While{$iter < I_{max}$}
    \State $[\mathbf{x}_0, \mathbf{x}_1,...,\mathbf{x}_T] \sim \mathcal{X}, [y_0, y_1,...,y_T] \sim \mathcal{Y}$ \hfill {\color{gray}\textit{// Sample $T$ trajectory windows}}
    \For{$t = 1$ to $T$} \hfill {\color{gray}\textit{// Number of rollouts}}
        \State $\mathbf{x}_t^0 \gets \mathbf{x}_t$
        \State $n \sim \mathcal{U}[0, N)$
        \State $\mathbf{x}_t^n \gets \textsc{Forward\_diffusion}(\mathbf{x}_t^0, n)$
        \State $\hat{\mathbf{x}}_t^0 = \mathcal{G}_{\theta}(\mathbf{x}_t^n, \mathbf{x}_{t-1},t, y_t)$ \hfill {\color{gray}\textit{// Next state denoising}}
        \State $\nabla \gets \nabla_{\theta} \mathcal{L}(\mathbf{x}_t^0, \hat{\mathbf{x}}_t^0)$
        \State $ \theta \gets \mathcal{O}(\theta, \nabla)$ \hfill {\color{gray}\textit{// Back propagation}}
        \State $p \gets$ \textsc{Schedule\_probability}($iter$, $I_{1}$, $I_2$) \hfill {\color{gray}\textit{// Sample schedule prob}}
        \If{$\text{rand}() < p$}
            \State $\mathbf{x}_t^N \gets \textsc{Forward\_diffusion}(\mathbf{x}_t^0, N)$ \hfill {\color{gray}\textit{// Maximum noising}}
            \State $\tilde{\mathbf{x}}_t^0 = \mathcal{S}(\mathcal{G}_{\theta}, \mathbf{x}_t^N,  \mathbf{x}_{t-1}, N, y_t)$ \hfill {\color{gray}\textit{// Full trajectory sampling}}
            \State $\mathbf{x}_t \gets \tilde{\mathbf{x}}_t^0$ \hfill {\color{gray}\textit{// Set predicted state into history}}
        \Else
            \State $\mathbf{x}_t \gets \mathbf{x}_t^0$ \hfill {\color{gray}\textit{// Use dataset state history}}
        \EndIf
        \State $iter \gets iter + 1$
    \EndFor
\EndWhile
\end{algorithmic}
\end{algorithm*}

\paragraph{Implementation Details of Diffusion Models.}
The denoiser $\mathcal{G}$ is implemented on the DIT backbone with 8 Transformer decoder layers (hidden size 512). We apply classifier‑free guidance by masking the terrain embedding with probability 0.1 during both training and inference. Local terrain is encoded from a 187-egocentric height map using lightweight MLP blocks. The final feature map is flattened and projected to a 256‑dimensional terrain embedding.

We train for a total of 300K diffusion iterations on our expert locomotion dataset, selecting the best checkpoints at 250K for validation. Optimization is via Adam with batch size 1024. All models run on a single NVIDIA GeForce RTX 4090 GPU for nearly 30 hours. At test time, we use DDPM sampling with guidance scale $\omega = 5$ to autoregressively generate continuous trajectories in real time.

\subsection{Details of Goal-conditioned Unified Policy}
\label{sup:unified}

\paragraph{Model Architecture.}

The network architecture is inherited from Section~\ref{sup:specialist}. Different from a specialized policy network, the goal-conditioned policy network contains an RNN future encoder $\mathcal{E}_\text{future}$, which encodes a predicted future state sequence $g_t = [\hat{s}_{t+1}, \ldots, \hat{s}_{t+H}] \in \mathbb{R}^{d \times H}$ sampled from the diffusion world model. The predicted future states are represented in the same body-relative parameterization used by the diffusion and reward function. The architecture of the unified policy is formulated as
\begin{equation}
    a_t = \mathcal{F} \left( \mathcal{E}_\text{hist}(o_t^{\text{hist}}),\; o_t^\text{proprio},\; \mathcal{E}_\text{future}(g_t),\; c_t,\; a_{t-1} \right).
    \label{eqa:unified_policy_archi}
\end{equation}

\paragraph{Reward formulations.}

Different from the reward functions used during specialized-policy training, the unified policy is trained with an immediate body-space tracking reward together with a long-horizon adversarial reward. Following BeyondMimic~\cite{liao2025beyondmimic}, we adopt a unified task-space body-tracking formulation for the immediate reward. Specifically, let $\mathcal{B}_{\mathrm{target}}$ denote the set of tracked bodies, and let the desired body poses and twists be extracted from the first predicted step $\hat{s}_{t+1}$ of the diffusion world model. In practice, these desired quantities can be read directly from $\hat{s}_{t+1}$ or computed from it via forward kinematics in the body-relative frame.

\begin{table}[t]
    \centering
    \caption{Immediate Body-space Tracking Reward for Unified Policy.}
    \begin{tabular}{lcc}
    \toprule
    \textbf{Tracking Term} & \textbf{Score} & \textbf{Tolerance} \\
    \midrule
    \textbf{Position} & $\exp(-\bar{e}_{p}/\sigma_{p}^{2})$ & $\sigma_{p}$ \\
    \textbf{Orientation} & $\exp(-\bar{e}_{R}/\sigma_{R}^{2})$ & $\sigma_{R}$ \\
    \textbf{Linear Velocity} & $\exp(-\bar{e}_{v}/\sigma_{v}^{2})$ & $\sigma_{v}$ \\
    \textbf{Angular Velocity} & $\exp(-\bar{e}_{\omega}/\sigma_{\omega}^{2})$ & $\sigma_{\omega}$ \\
    \midrule
    \textbf{Total} & $\sum_{s \in \{\mathbf{p}, R, \mathbf{v}, \boldsymbol{\omega}\}} \exp(-\bar{e}_{s}/\sigma_{s}^{2})$ & -- \\
    \bottomrule
    \end{tabular}
    \label{tab:supp_unified_reward}
\end{table}

For each target body $b \in \mathcal{B}_{\mathrm{target}}$, we define
\begin{equation}
\mathbf{e}_{p,b} = \mathbf{p}^{\mathrm{des}}_b - \mathbf{p}_b,\qquad
\mathbf{e}_{R,b} = \log\left(R_b^{\mathrm{des}} R_b^\top\right),
\end{equation}
\begin{equation}
\mathbf{e}_{v,b} = \mathbf{v}^{\mathrm{des}}_b - \mathbf{v}_b,\qquad
\mathbf{e}_{\omega,b} = \boldsymbol{\omega}^{\mathrm{des}}_b - \boldsymbol{\omega}_b.
\end{equation}
The mean squared error of each tracking term is averaged over all target bodies:
\begin{equation}
\bar{e}_s
=
\frac{1}{|\mathcal{B}_{\mathrm{target}}|}
\sum_{b \in \mathcal{B}_{\mathrm{target}}}
\left\| \mathbf{e}_{s,b} \right\|^2,
\qquad
s \in \{\mathbf{p}, R, \mathbf{v}, \boldsymbol{\omega}\}.
\end{equation}
Each term is then mapped through a Gaussian-shaped score:
\begin{equation}
r(\bar{e}_s, \sigma_s)
=
\exp\left(-\bar{e}_s/\sigma_s^2\right),
\end{equation}
and the immediate tracking reward is defined as
\begin{equation}
r_t^\text{track}
=
\sum_{s \in \{\mathbf{p}, R, \mathbf{v}, \boldsymbol{\omega}\}}
r(\bar{e}_s, \sigma_s).
\label{eqa:unified_track_reward}
\end{equation}

To complement the body-tracking objective, we introduce a modified AMP-style~\cite{peng2021amp} adversarial reward. Different from standard AMP, the discriminator is updated using future motion plans online-sampled from the diffusion world model rather than pre-recorded motion clips, as described in Algorithm~\ref{alg:supp_amp_diffusion}. Therefore, $r_t^\text{track}$ provides immediate body-level geometric supervision, while $r_t^\text{amp}$ encourages temporally consistent and terrain-adaptive whole-body motion.

\begin{algorithm*}[t]
\caption{Training with Online Diffusion AMP}
\label{alg:supp_amp_diffusion}
\begin{algorithmic}[1]
\State \textbf{Input} $\mathcal{M}_{\text{diff}}$: autoregressive diffusion world model, $K$: history length, $H$: horizon
\State initialize discriminator $D$, policy $\pi$, value function $V$, replay buffer $\mathcal{B} \leftarrow \emptyset$, history buffer $\mathcal{H} \leftarrow \emptyset$.
\While{not done}
    \For{trajectory $i=1,...,m$} \hfill {\color{gray}\textit{// Data collection phase}}
        \State $\tau^{i} \leftarrow \emptyset$
        \For{step $t=0,\ldots,T-1$}
            \State Execute $a_t \sim \pi(s_t)$, observe $s_{t+1}$, store $(s_t,a_t,r_t^\text{regu},s_{t+1})$ in $\mathcal{B}$
            \State Update $\mathcal{H}$ with $s_t$: $\mathcal{H} \leftarrow \text{concat}(\mathcal{H}[-K+1:], s_t)$ \hfill {\color{gray}\textit{// Maintain latest $K$ states}}
            \If{$t < K-1$} \hfill {\color{gray}\textit{// Padding for initial phase}}
                \State $\hat{\mathcal{H}} \leftarrow [s_0,\ ...,\ s_0, \mathcal{H}[0:t+1]]$ \hfill {\color{gray}\textit{// Pad with $s_0$}}
            \Else
                \State $\hat{\mathcal{H}} \leftarrow \mathcal{H}[-K:]$ \hfill {\color{gray}\textit{// Last $K$ states}}
            \EndIf
            \State Generate $\{\hat{s}_{t+1:t+H}\} \sim \mathcal{M}_{\text{diff}}(\hat{\mathcal{H}})$
            \State $d_t \gets D(\Phi(s_t),\Phi(\hat{s}_{t+1}))$ \hfill {\color{gray}\textit{// Use immediate prediction}}
            \State $r_t^\text{amp} \leftarrow$ style reward via Eq.\ref{eqa:amp-reward-main} using $d_t$
            \State $r_t \leftarrow w^\text{regu} r_t^\text{regu} + w^\text{track} r_t^\text{track} + w^\text{amp} r_t^\text{amp}$
            \State Append $(s_t,a_t,r_t,s_{t+1})$ to $\tau^i$
        \EndFor
    \EndFor
    \For{update step $=1,...,n$} \hfill {\color{gray}\textit{// Training phase}}
        \State $b^{\pi} \leftarrow$ sample $K$ transitions $\{(s_j,s_j')\}_{j=1}^K$ from $\mathcal{B}$
        \State Generate $\{\hat{s}_{j:j+H}\}_{j=1}^K \sim \mathcal{M}_{\text{diff}}(b^{\pi})$ \hfill {\color{gray}\textit{// On-the-fly generation}}
        \State Update $D$ via Eq.\ref{eqa:reward_discriminator} using $\{(\Phi(s_j),\Phi(\hat{s}_{j+1}))\}_{j=1}^K$ \hfill {\color{gray}\textit{// Fresh samples}}
        \State Update $V$ and $\pi$ using $\{\tau^i\}_{i=1}^m$ \hfill {\color{gray}\textit{// Policy optimization}}
    \EndFor
\EndWhile
\end{algorithmic}
\end{algorithm*}

\paragraph{Reward Functions and Weight.}

Table~\ref{tab:supp_unified_reward_weight} lists the effective scalar reward weights used for unified policy learning. 
The term "effective weight" denotes the coefficient directly applied to each scalar reward before summation in Eq.~\ref{eqa:unified_total_reward}. 
The reward design contains 19 scalar terms in total: 4 body-space tracking terms, 10 physical regularization terms, 4 safety terms, and 1 diffusion-AMP style term. 
Compared with the specialized policies, we remove terrain-specific shaping rewards and keep only embodiment-level regularization, because terrain-dependent motion targets are provided by the diffusion future trajectory $g_t=\hat{s}_{t+1:t+H}$.

\begin{table*}[t]
\centering
\caption{
\textbf{Reward Functions and Weights for the Unified Policy.}
Notation: $\mathcal{B}_{\mathrm{target}}$ denotes the tracked body set; 
$q$, $\dot{q}$, $\ddot{q}$ denote joint position, velocity, and acceleration; 
$\tau$ denotes joint torque; 
$a_t$ denotes policy action; 
$c_f$ is the contact indicator of foot $f$; 
$\mathbf{g}^{\mathrm{proj}}$ is the projected gravity vector; 
$\mathbb{I}(\cdot)$ is the indicator function.
}
\label{tab:supp_unified_reward_weight}
\small
\begin{tabular}{llc}
\toprule
\textbf{Reward Function} & \textbf{Formula} & \textbf{Weight} \\
\midrule
\multicolumn{3}{l}{\textit{Diffusion-guided Tracking Reward}} \\
Body Position Tracking &
$\exp\left(-\bar{e}_{\mathbf{p}}/\sigma_{\mathbf{p}}^2\right)$ &
$1.0$ \\
Body Orientation Tracking &
$\exp\left(-\bar{e}_{R}/\sigma_{R}^2\right)$ &
$1.0$ \\
Body Linear Velocity Tracking &
$\exp\left(-\bar{e}_{\mathbf{v}}/\sigma_{\mathbf{v}}^2\right)$ &
$1.0$ \\
Body Angular Velocity Tracking &
$\exp\left(-\bar{e}_{\boldsymbol{\omega}}/\sigma_{\boldsymbol{\omega}}^2\right)$ &
$1.0$ \\
\midrule
\multicolumn{3}{l}{\textit{Physical Regularization Reward}} \\
Linear Velocity Z &
$v_z^2$ &
$-1.0$ \\
Base Angular Velocity XY &
$\|\boldsymbol{\omega}_{xy}\|_2^2$ &
$-0.05$ \\
Hip Joint Deviation &
$\sum_{j\in\mathcal{J}_{\mathrm{hip}}}(q_j-q_{j,\mathrm{default}})^2$ &
$-0.5$ \\
Joint Torques $(L_2)$ &
$\|\boldsymbol{\tau}\|_2^2$ &
$-1.5\mathrm{e}{-7}$ \\
Joint Acceleration $(L_2)$ &
$\|\ddot{\mathbf{q}}\|_2^2$ &
$-1.25\mathrm{e}{-7}$ \\
Joint Velocity $(L_2)$ &
$\|\dot{\mathbf{q}}\|_2^2$ &
$-1.0\mathrm{e}{-4}$ \\
Action Rate $(L_2)$ &
$\|a_t-a_{t-1}\|_2^2$ &
$-0.005$ \\
Flat Orientation &
$\|\mathbf{g}^{\mathrm{proj}}_{xy}\|_2^2$ &
$-3.0$ \\
Feet Slide &
$\sum_f \|\mathbf{v}_{xy,f}\|_2 \cdot \mathbb{I}(c_f)$ &
$-0.4$ \\
Energy Consumption &
$\sum_j(\tau_j\dot{q}_j/k_j)^2$ &
$-5.0\mathrm{e}{-5}$ \\
\midrule
\multicolumn{3}{l}{\textit{Safety Reward}} \\
Joint Position Limits &
$\sum_j\left(\max(0,q_j-q_{j,\max})+\max(0,q_{j,\min}-q_j)\right)$ &
$-1.0$ \\
Joint Velocity Limits &
$\sum_j\max(0,|\dot{q}_j|-0.9\dot{q}_{j,\max})$ &
$-1.0$ \\
Torque Limits &
$\sum_j\max(0,|\tau_j|-0.8\tau_{j,\max})^2$ &
$-0.01$ \\
Undesired Contacts &
$\mathbb{I}\left(\mathrm{count}(\mathrm{collision}_{\mathrm{body}\setminus\mathrm{feet}})>0\right)$ &
$-1.0$ \\
\midrule
\multicolumn{3}{l}{\textit{Diffusion-AMP Style Reward}} \\
AMP Style &
$\max\left\{0,\ 1-0.25\left(D(\Phi(s_t),\Phi(s_{t+1}))-1\right)^2\right\}$ &
$0.25$ \\
\bottomrule
\end{tabular}
\end{table*}

For the tracking terms, the body-space errors follow the definitions in Eq.~\ref{eqa:unified_track_reward}. 
Specifically, the desired body pose and twist are extracted from the first predicted state $\hat{s}_{t+1}$ generated by the diffusion world model. 
The four tracking terms provide immediate geometric supervision, while the diffusion-AMP term in Eq.~\ref{eqa:amp-reward-main} regularizes the policy toward temporally consistent motion patterns over the generated future trajectory. 
Therefore, the unified reward remains compact while still capturing terrain-adaptive whole-body locomotion behavior.

\paragraph{DreamPolicy Inference Process.}
The inference process is addressed in Algorithm \ref{alg:supp_inference}: the agent first observes the current state \(s_t\) from the environment and acquires the terrain heightmap, stores the current state in the history buffer, then interacts with the environment to obtain the action, executes the action, and finally transitions to the next state \(s_\text{t+1}\).

\begin{algorithm*}[t]
\caption{DreamPolicy Inference}
\label{alg:supp_inference}
\begin{algorithmic}[1]
\State \textbf{Input} $\mathcal{M}_{\text{diff}}$: trained diffusion world model, $\pi$: trained policy, $K$: history length, $H$: horizon
\State $\mathcal{H} \leftarrow \emptyset$ initialize history buffer
\State $z_{\text{terrain}} \leftarrow$ preprocessed terrain heightmap
\While{running}
    \State Observe current state $s_t$
    \State Update $\mathcal{H}$: $\mathcal{H} \leftarrow \text{concat}(\mathcal{H}[-K+1:], s_t)$ \hfill {\color{gray}\textit{// Maintain latest $K$ states}}
    \If{$|\mathcal{H}| < K$} \hfill {\color{gray}\textit{// Initial phase padding}}
        \State $\hat{\mathcal{H}} \leftarrow [s_0,\ ...,\ s_0, \mathcal{H}[0:t+1]]$ \hfill {\color{gray}\textit{// Pad with initial state}}
    \Else
        \State $\hat{\mathcal{H}} \leftarrow \mathcal{H}[-K:]$ \hfill {\color{gray}\textit{// Last $K$ states}}
    \EndIf
    \State Generate $\{\hat{s}_{t+1:t+H}\} \sim \mathcal{M}_{\text{diff}}(\hat{\mathcal{H}}, z_{\text{terrain}})$ \hfill {\color{gray}\textit{// Terrain-conditioned prediction}}
    \State Construct policy input: $\tilde{s}_t \leftarrow [s_t; \hat{s}_{t+1:t+H}]$ \hfill {\color{gray}\textit{// Concatenate future predictions}}
    \State $a_t \leftarrow \pi(\tilde{s}_t)$ \hfill {\color{gray}\textit{// Generate action}}
    \State Execute $a_t$ on robot and observe next state $s_{t+1}$
\EndWhile
\end{algorithmic}
\end{algorithm*}

\section{Additional Experiments and Results}
\label{sup:result}

\subsection{Detailed Implementation of Baselines}
\label{sup:baseline}

\paragraph{Scenario Specialized Policy.}
Inspired by \cite{cheng2024extreme}, we use a scandot encoder and a privileged information estimator, as mentioned in Section \ref{sup:specialist}. To this end, we will not use this setting for experiments on multi-terrains.

\paragraph{Distillation-based Method.}
Inspired by \cite{zhuang2023robot}, which trained a unified policy using online distillation via DAgger \cite{ross2011reduction}. We trained a policy via online distillation, where the unified policy learn from multiple single specialized policies via supervision.

\paragraph{Ablation on Diffusion World Model.}
In this setting, we remove the autoregressive diffusion world model. Due to the lack of motion guidance, we train the discriminator using the standard implementation of AMP \cite{peng2021amp}, in which we sample transitions from offline trajectories collected from specialized policies.

\paragraph{Ablation on Goal-conditioned RL.}
In this setting, we use the action output by the autoregressive diffusion world model directly rather than state predictions for RL guidance. This setting aims to demonstrate that diffusion directly predicts actions with poor performance, as illustrated in \ref{sup:experiments}; therefore, using prediction and reinforcement learning training on more continuous state spaces can achieve dynamically feasible multi-terrain adaptation.

\subsection{Additional Experiments}
\label{sup:experiments}

\begin{figure}[b]
  \centering
  \includegraphics[width=1.0\textwidth]{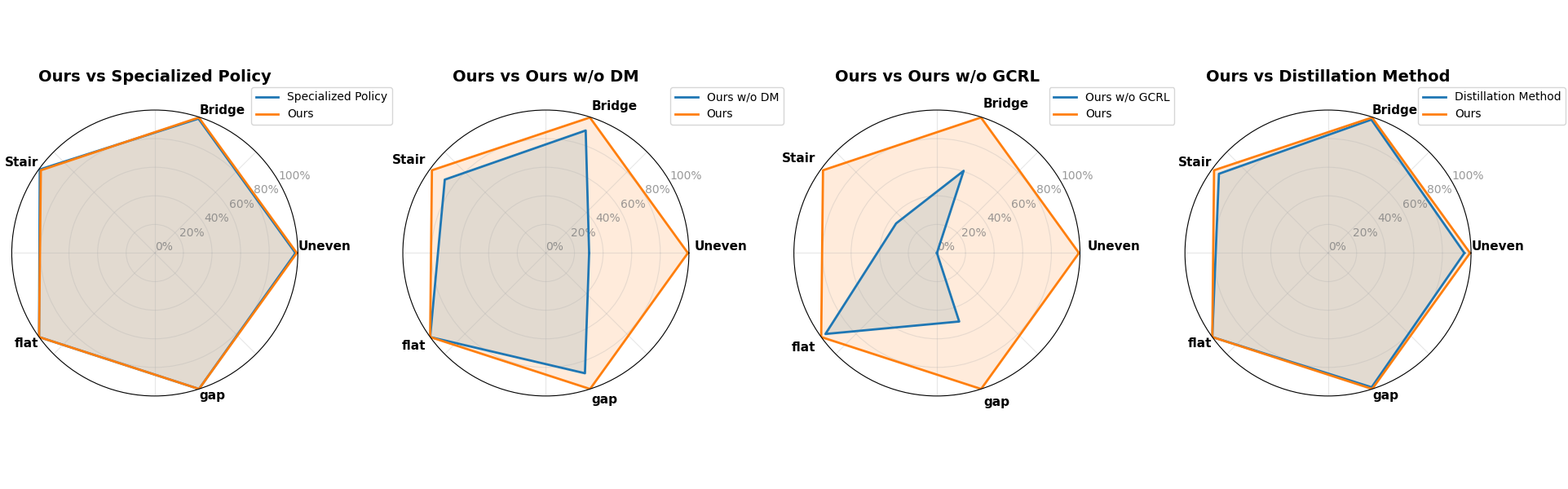}
  \caption{\textbf{Comparison of success rates across single terrains.} This figure compares the DreamPolicy against four baselines across five single terrain types. Each subplot contrasts Ours with one baseline method, with radial axes representing success rates.}
  \label{exp:single_terrain_figure}
\end{figure}

Table \ref{add:single_terrain} shows the detailed data on a single terrain that we display using lidar maps in the main text. The detailed data shows that our method does not exhibit significant performance degradation. 

\setlength{\tabcolsep}{4pt}
\begin{table*}[!ht]
\caption{\textbf{Settings Comparison in Single-Terrain}. All data in the table are shown as percentages; the higher the better. The best results are highlighted in boldface.}
\begin{center}
 \resizebox{\linewidth}{!}{
\begin{tabular}{lcccccccccc}
\toprule[1.0pt]
\multicolumn{1}{l}{\multirow{2}{*}{}} & \multicolumn{2}{c}{Uneven} & \multicolumn{2}{c}{Bridge} & \multicolumn{2}{c}{Stair} &
\multicolumn{2}{c}{Flat} &
\multicolumn{2}{c}{Gap}\\
\cmidrule[\heavyrulewidth](lr){2-3} \cmidrule[\heavyrulewidth](lr){4-5} \cmidrule[\heavyrulewidth](lr){6-7} \cmidrule[\heavyrulewidth](lr){8-9}\cmidrule[\heavyrulewidth](lr){10-11}

\multicolumn{1}{l}{} & $R_{\mathrm{succ}}$ ($\%$) & $R_\mathrm{cmp}$  ($\%$)& $R_\mathrm{succ}$ ($\%$) & $R_\mathrm{cmp}$ ($\%$) & $R_{\mathrm{succ}}$ ($\%$) & $R_{\mathrm{cmp}}$ ($\%$) & $R_\mathrm{succ}$  ($\%$)& $R_{\mathrm{cmp}}$ ($\%$) & $R_\mathrm{succ}$  ($\%$) & $R_\mathrm{cmp}$  ($\%$) \\
\midrule[0.8pt]

Specialized Policy  & $98.34$ & $99.16$ & $98.82$ & $98.78$ & \textbf{99.42} & \textbf{99.23}  & \textbf{100.00} & \textbf{100.00}  & \textbf{100.00} & \textbf{100.00}\\ [0.4ex]

Distillation  & $95.33$ & $91.21$ & $98.31$ & \textbf{99.36} & $94.65$ & $95.22$ & \textbf{100.00} & \textbf{100.00} & $98.78$ & $96.54$\\  [0.4ex]

Ours  & \textbf{99.37} & \textbf{99.22} & \textbf{99.65} & $99.17$ & $98.46$ & $97.13$  & \textbf{100.00} & \textbf{100.00} & \textbf{100.00} & \textbf{100.0} \\  [0.4ex]
\bottomrule[1.0pt]
Ours w/o Diffusion  & $30.25$ & $41.45$ & $90.11$ & $92.52$ & $87.64$ & $78.44$ & \textbf{100.00} & \textbf{100.00}  & $88.53$ & $86.87$ \\  [0.4ex]

Ours w/o GCRL  & $0.00$ & $1.27$ & $60.32$ & $54.39$ & $35.08$ & $22.18$  & $96.34$ & $98.24$ & $50.86$ & $62.37$ \\  [0.4ex]

\bottomrule[1.0pt]
\end{tabular}
}
\end{center}
\end{table*}
\label{add:single_terrain}

\subsection{Primitive Terrain Analysis} \label{sup:visual}

\begin{figure}[t]
  \centering
  \includegraphics[width=0.6\textwidth]{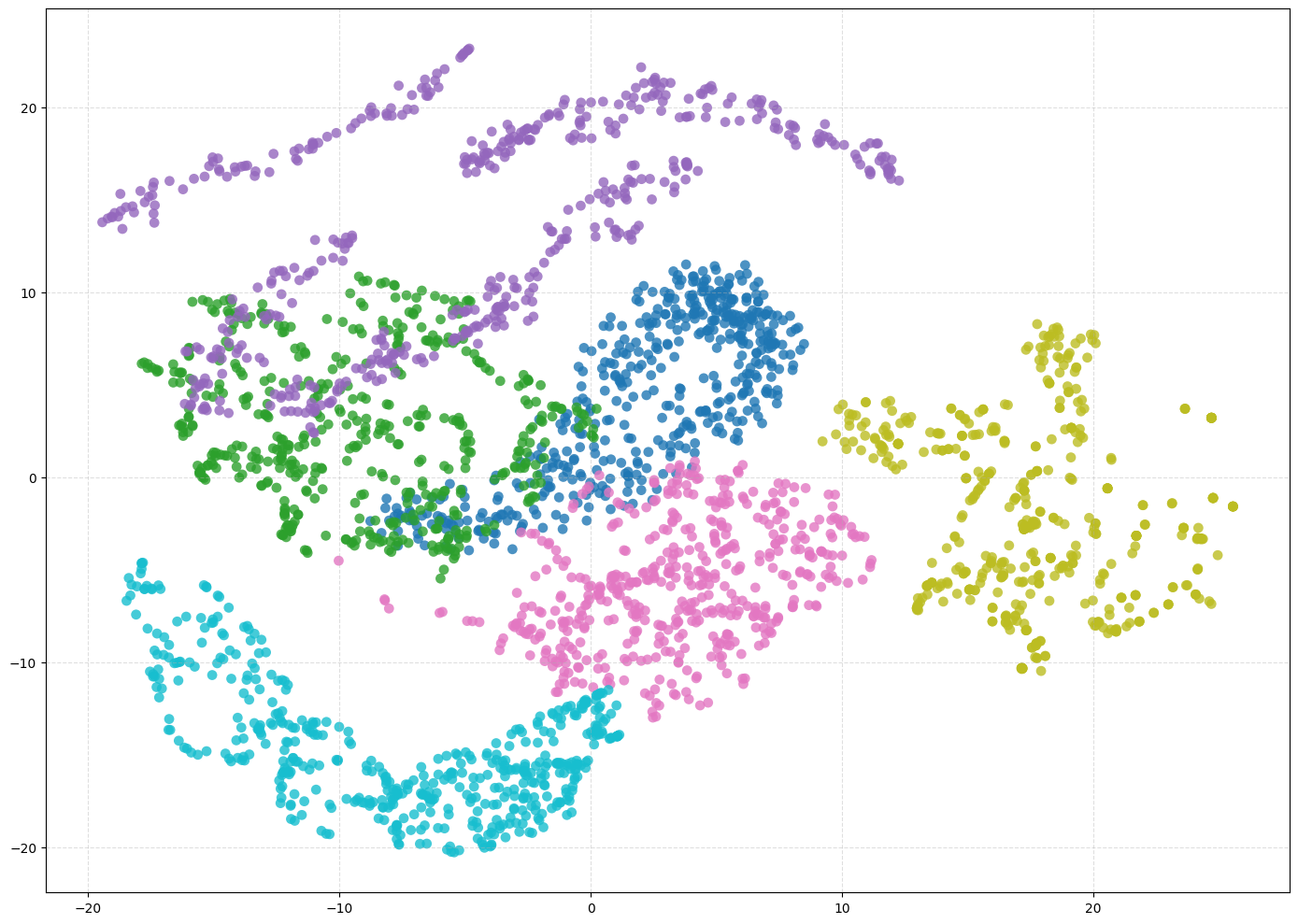}
  \caption{Terrain t-SNE visualization.}
  \label{exp:tsne}
\end{figure}

The t-SNE visualization (Fig.~\ref{exp:tsne}) reveals a structured latent space where the \textbf{Plane} (Blue) acts as a central hub, with complex terrains forming distinct, specialized clusters. Kinematically, these clusters represent conflicting motion modes: \textbf{Stairs} (Purple) demand high hip flexion for clearance, while \textbf{Stepping Stones} (Yellow) require precise, discrete foot placement. 

Conventional distillation fails here by collapsing these multi-modal distributions into a mean representation, causing instability in mixed scenarios. In contrast, our framework treats these clusters as composable primitives. By learning to smoothly interpolate between these distinct kinematic modes rather than averaging them, our method enables robust, zero-shot adaptation to unseen composite terrains.

\begin{figure*}[t]
    \centering

    \begin{subfigure}[b]{0.51\linewidth}
        \centering
        \includegraphics[width=\linewidth]{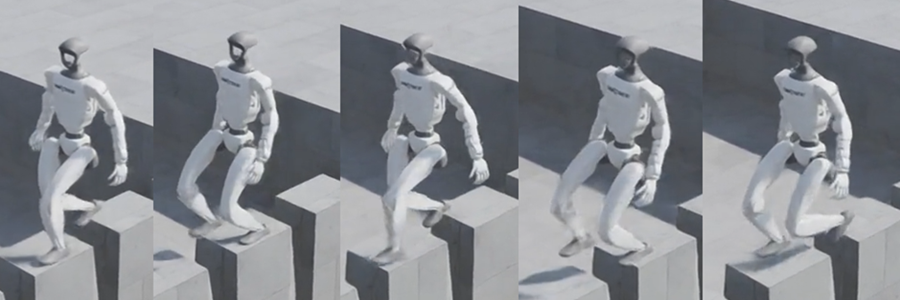}
        \caption{Simulation qualitative Results on Balancing Beam.}
        \label{fig:gap_bridge}
    \end{subfigure}
    \hfill
    \begin{subfigure}[b]{0.47\linewidth}
        \centering
        \includegraphics[width=\linewidth]{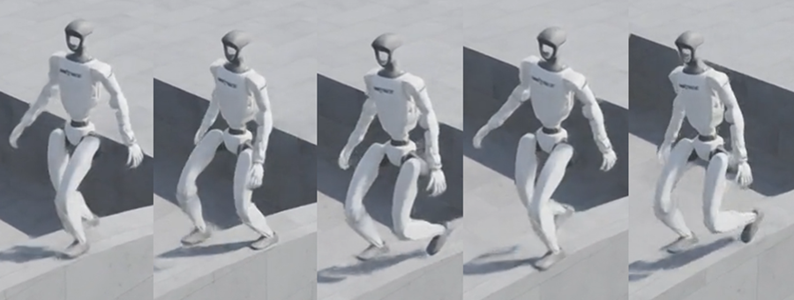}
        \caption{Simulation qualitative result on Slope-Bridge.}
        \label{fig:slope_bridge}
    \end{subfigure}

    \vspace{0.5em}

    \begin{subfigure}[b]{0.51\linewidth}
        \centering
        \includegraphics[width=\linewidth]{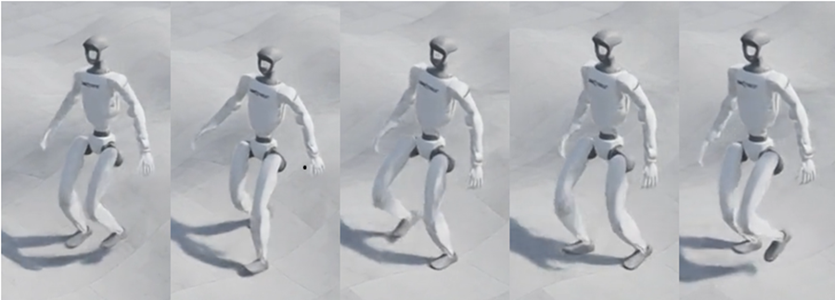}
        \caption{Simulation qualitative result on Wave Terrain.}
        \label{fig:wave}
    \end{subfigure}
    \hfill
    \begin{subfigure}[b]{0.47\linewidth}
        \centering
        \includegraphics[width=\linewidth]{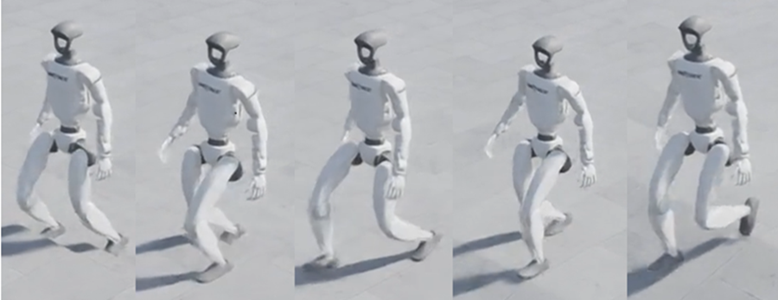}
        \caption{Simulation qualitative result on multi-terrain.}
        \label{fig:multi}
    \end{subfigure}
    \caption{Simulation qualitative result in IsaacLab.}
    \label{fig:sim-result-appendix1}
\end{figure*}

More simulation results is shown in Figure \ref{fig:sim-result-appendix1}, \ref{fig:sim-result-appendix1}

\begin{figure}[t]
    \centering
    \begin{subfigure}{0.49\linewidth}
        \centering
        \includegraphics[width=\linewidth]{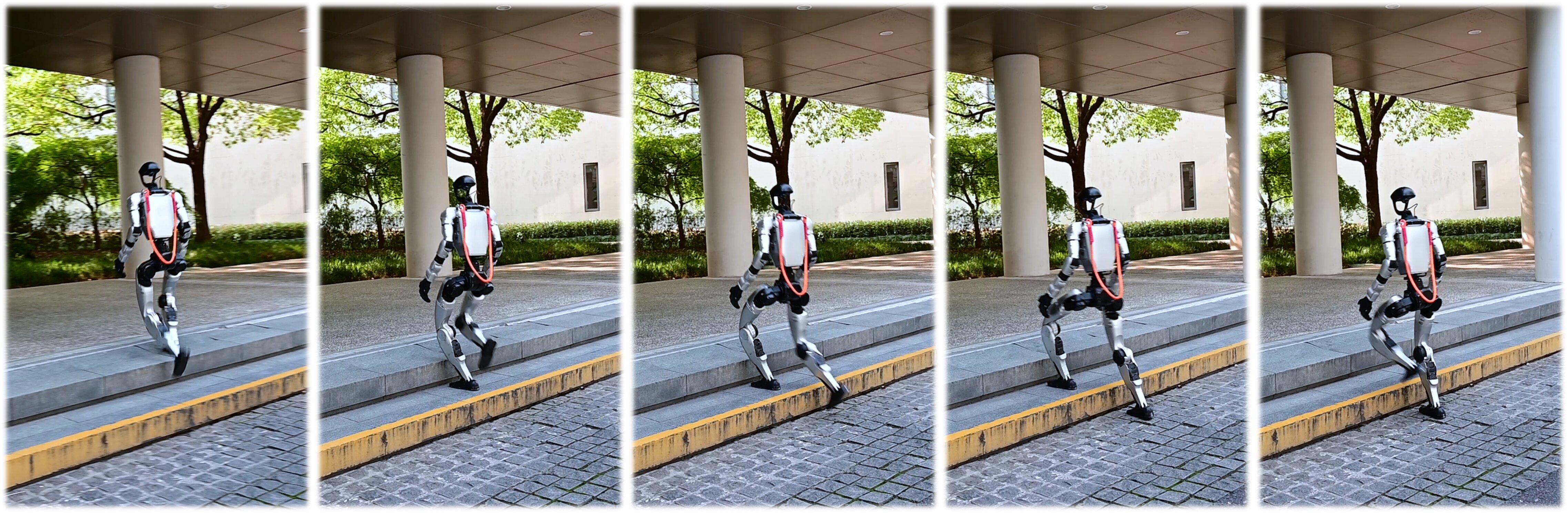}
        \caption{Outdoor qualitative results for upstairs scenario.}
        \label{fig:appendix-real-1}
    \end{subfigure}
    \hfill
    \begin{subfigure}{0.47\linewidth}
        \centering
        \includegraphics[width=\linewidth]{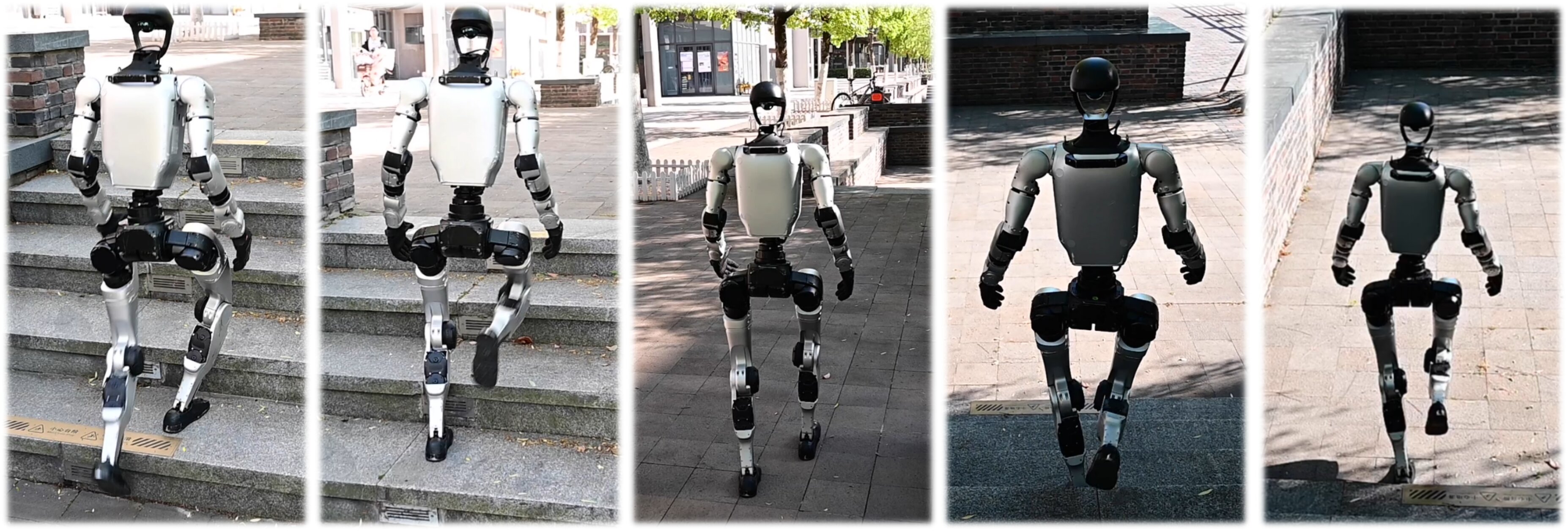}
        \caption{Outdoor qualitative results on upstairs scenario.}
        \label{fig:appendix-real-2}
    \end{subfigure}
    \vspace{0.5em}
    \begin{subfigure}{0.48\linewidth}
        \centering
        \includegraphics[width=\linewidth]{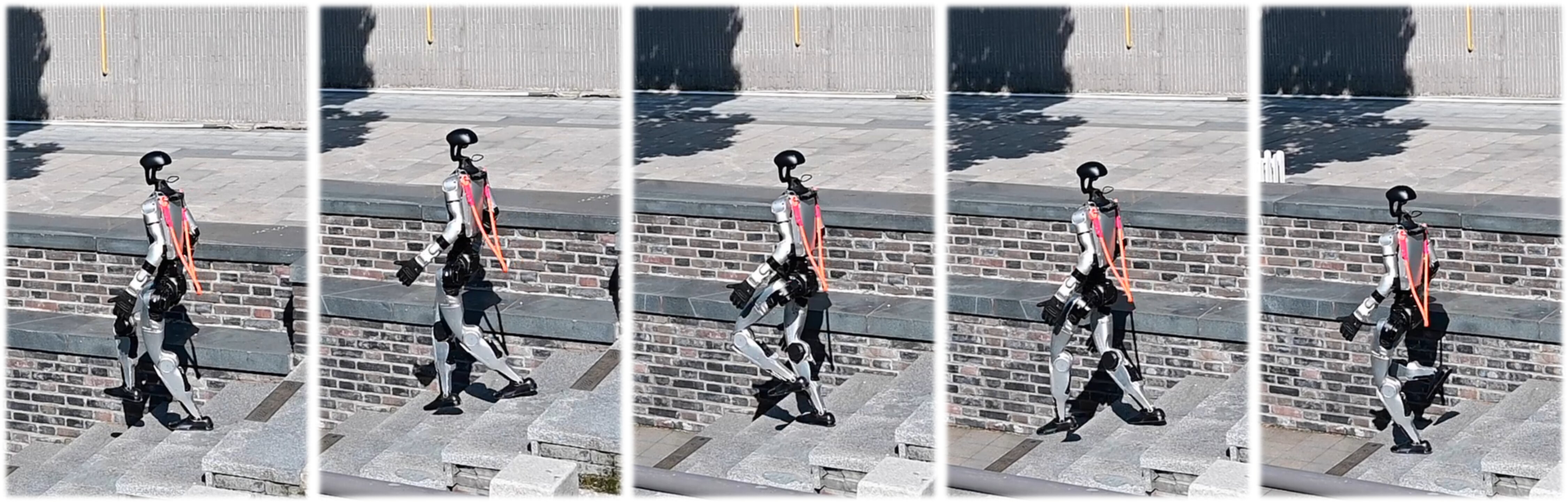}
        \caption{Outdoor qualitative results on downstairs scenario.}
        \label{fig:appendix-real-3}
    \end{subfigure}
    \hfill
    \begin{subfigure}{0.48\linewidth}
        \centering
        \includegraphics[width=\linewidth]{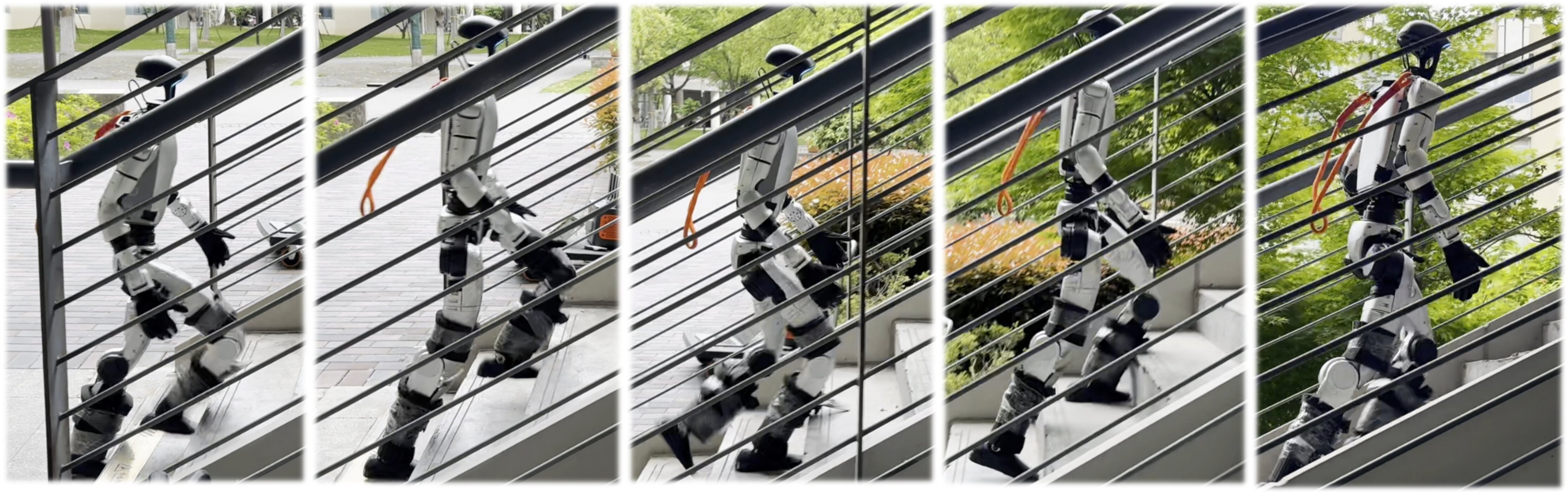}
        \caption{Outdoor qualitative results on long upstairs slope.}
        \label{fig:appendix-real-8}
    \end{subfigure}

    \vspace{0.5em}
    \begin{subfigure}{0.48\linewidth}
        \centering
        \includegraphics[width=\linewidth]{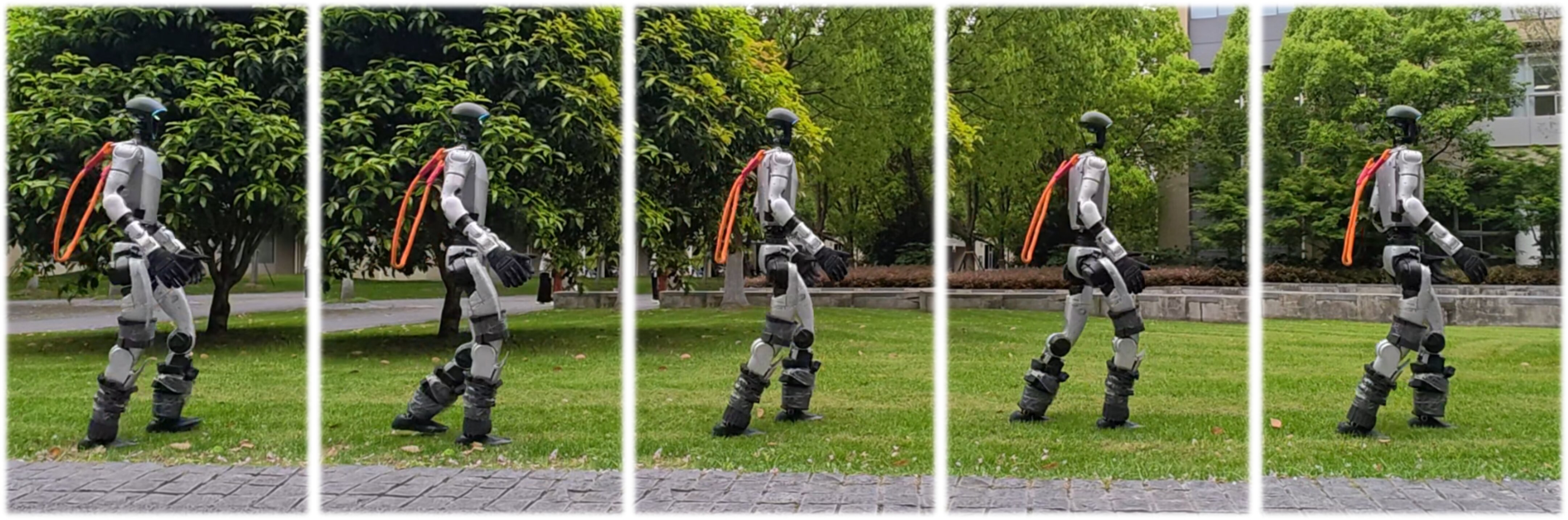}
        \caption{Outdoor qualitative results on grass scenario.}
        \label{fig:appendix-real-7}
    \end{subfigure}
    \hfill
    \begin{subfigure}{0.48\linewidth}
        \centering
        \includegraphics[width=\linewidth]{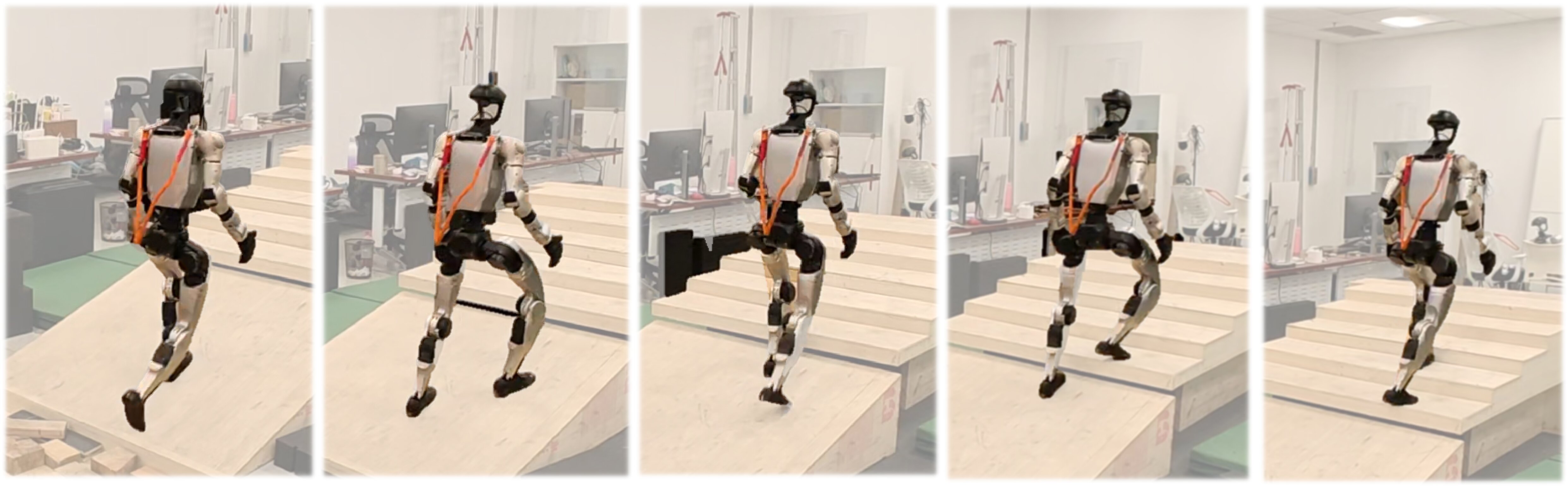}
        \caption{Indoor qualitative results on an upward slope.}
        \label{fig:appendix-real-4}
    \end{subfigure}
    \vspace{0.5em}
    \begin{subfigure}{0.48\linewidth}
        \centering
        \includegraphics[width=\linewidth]{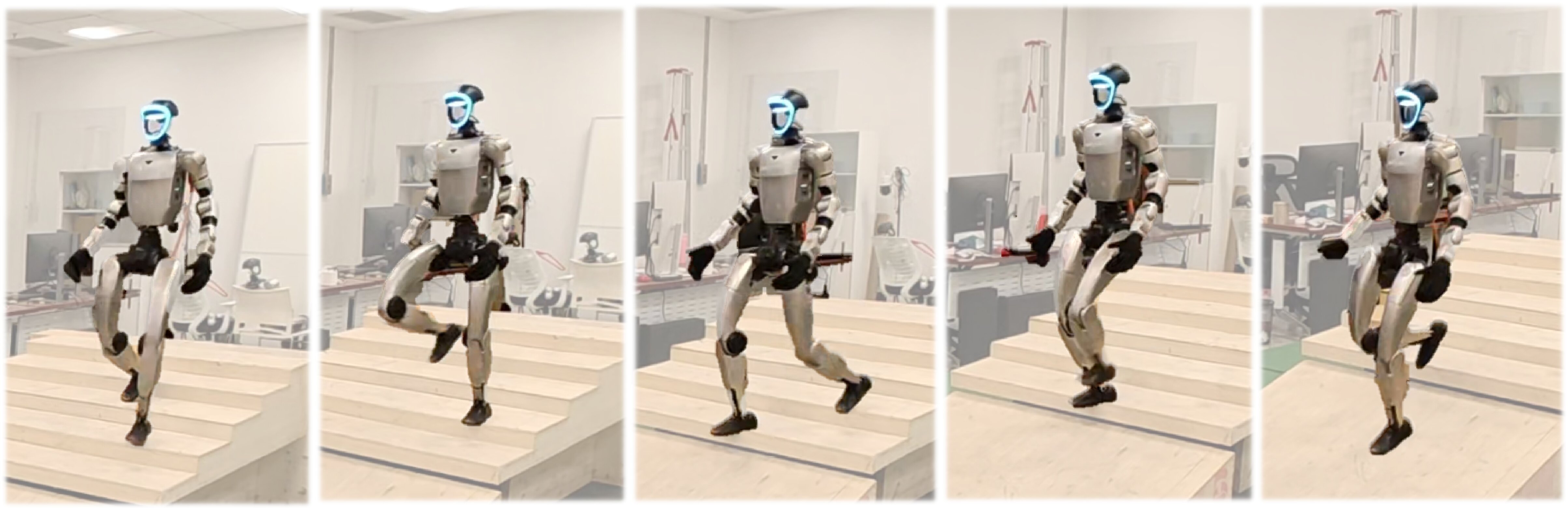}
        \caption{Indoor qualitative results for downstairs scenario.}
        \label{fig:appendix-real-5}
    \end{subfigure}
    \hfill
    \begin{subfigure}{0.48\linewidth}
        \centering
        \includegraphics[width=\linewidth]{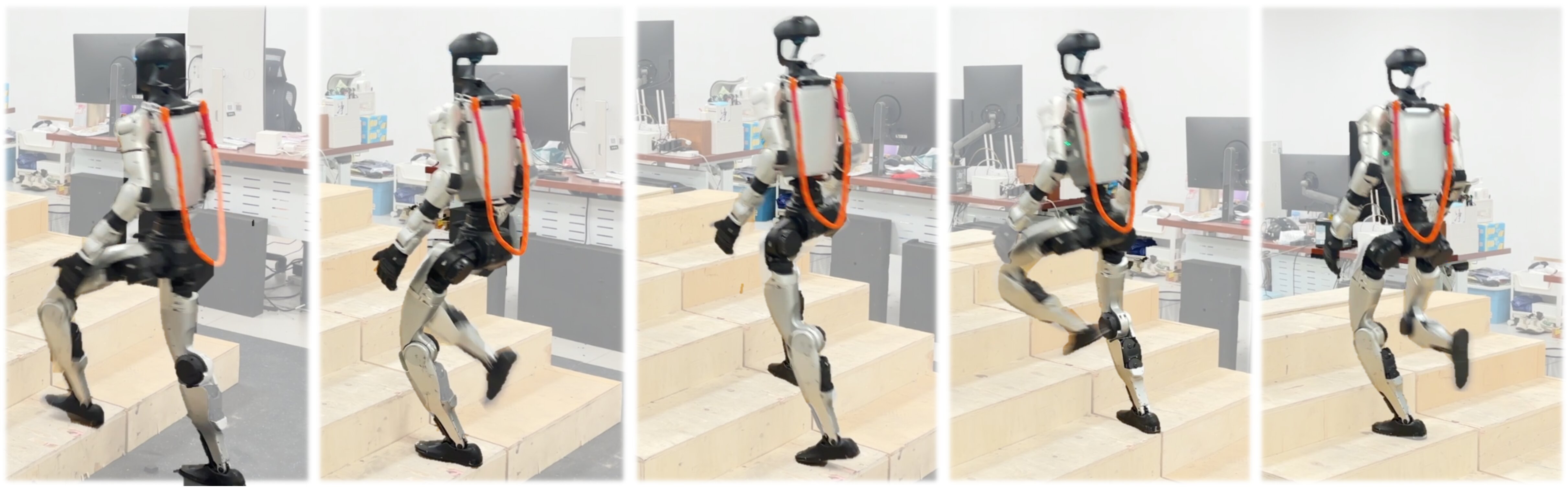}
        \caption{Indoor qualitative results for upstairs scenario.}
        \label{fig:appendix-real-6}
    \end{subfigure}
    \caption{Qualitative results from real-world demonstrations in outdoor and indoor environments.}
    \label{fig:appendix-real-demo}
\end{figure}

Additional real-world visualization is shown in Figure~\ref{fig:appendix-real-demo}.

\begin{figure*}[ht]
    \centering
    \includegraphics[width=0.95\linewidth]{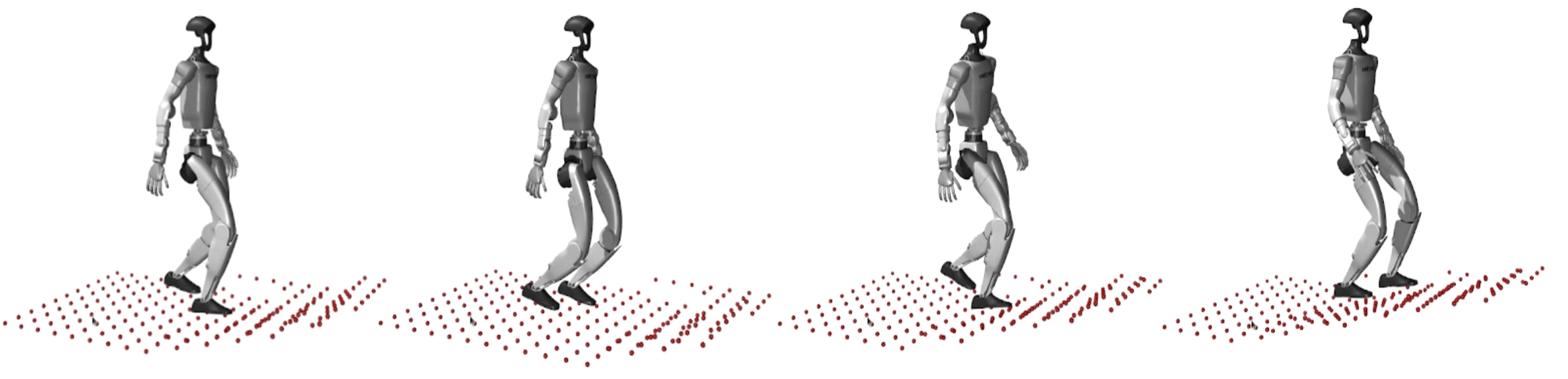}
    \caption{\textbf{Visualization of Terrain-Adaptive Motion.} The generated trajectories demonstrate the model's ability to proactively anticipate terrain changes, such as adjusting step height in advance for stairs.}
    \label{fig:composite_terrain}
\end{figure*}

\section{Supplementary Video}
\label{sup:video}
We have included a video in the supplementary material that summarizes our main contributions and presents video results of humanoid locomotion. More details are as follows:
\begin{itemize}
\item We introduced our DreamPolicy algorithmic framework, summarized our main contributions across four aspects, and detailed each component. 
\item We presented our quantitative results on single-terrain and unseen terrains and analyzed the impact of scale on our method.
\item We gave a visual comparison of our performance on different terrains with the traditional distillation method, showing that our algorithm has significant advantages.
\end{itemize}

\section{Extended Limitation and Discussions}
\label{sup:conclusion}

\subsection{Limitation and Failure Cases}
\label{sup:limitation}

Our analysis suggests that the limitations may stem from two factors:
1) \textbf{Naive terrain encoding.} The current diffusion world model employs a plain MLP to process the 187‑dimensional structured terrain elevation map, without any spatially aware components that could extract local geometric patterns and contextual cues essential for handling sharp gradients or complex surface discontinuities.  
2) \textbf{Limited policy perception.} Our unified policy relies exclusively on the planner’s state trajectory predictions for navigation, lacking any direct perception of the terrain. Consequently, the policy is unable to react to unforeseen geometric perturbations or dynamically adjust its foot placement in extreme, unstructured environments.  

These combined limitations—inadequate feature representation of challenging terrain and absence of real‑time sensory feedback—undermine stability and recovery capability. 

\subsection{Discussion and Future Work}
\label{sup:discussion}

We propose a set of comprehensive research directions and rigorous validation studies—spanning dataset augmentation, terrain‑generalization techniques, inference optimization, and real‑world robotic trials—to systematically extend the capabilities and verify the practical effectiveness of our diffusion‑based humanoid motion planning framework:

\begin{itemize}

\item \textbf{Augment dataset diversity.} Integrate additional data sources beyond the current set of expert policies trained in RL—including demonstrations from varied control algorithms, motion capture recordings of human subjects, and synthetic trajectories generated under different environmental perturbations—to significantly enrich the breadth, variability, and robustness of the humanoid motion imagery dataset.

\item \textbf{Enhance terrain generalization.} Develop and evaluate advanced terrain‑encoding strategies—such as learned geometric embeddings, height‑map spectral features, and multimodal descriptors—that can effectively generalize across a wide spectrum of previously unseen environments, thereby enabling the diffusion planner to produce reliable, accurate future‑state priors even on novel or complex terrains.

\item \textbf{Accelerate inference efficiency.} Reduce sampling latency by integrating techniques such as step‑reduction schedulers, diffusion model distillation into lightweight student networks, dynamic stopping criteria based on convergence metrics, or adaptive iteration strategies that allocate computation where it is most needed, all aimed at supporting true real‑time deployment on computationally constrained hardware.

\item \textbf{Standardize benchmarking.} Develop comprehensive, open evaluation suites—comprising diverse task scenarios, standardized test terrains, and reproducible performance metrics (e.g.\ success rate, energy consumption, and robustness measures)—to facilitate fair, transparent comparisons across diffusion‑based humanoid planners and to drive accelerated community progress.

\end{itemize}

\end{document}